\documentclass{article}

\usepackage{arxiv}

\usepackage[utf8]{inputenc} 
\usepackage[T1]{fontenc}    
\usepackage{hyperref}       
\usepackage{url}            
\usepackage{booktabs}       
\usepackage{amsfonts}       
\usepackage{nicefrac}       
\usepackage{microtype}      
\usepackage{lipsum}		
\usepackage{graphicx,psfrag,epsf}
\usepackage{natbib}
\usepackage{doi}
\usepackage{amsmath}
\usepackage{enumerate}
\usepackage{siunitx}
\usepackage{multirow}
\usepackage{amssymb}
\usepackage{appendix}
\usepackage[linesnumbered,ruled,vlined]{algorithm2e}
\usepackage[switch,pagewise]{lineno}

\title{Multi-Response Heteroscedastic Gaussian Process Models and Their Inference}

\author{ 
    \href{https://orcid.org/0000-0002-5936-762X}{\includegraphics[scale=0.06]{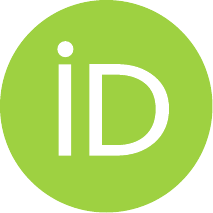}\hspace{1mm}Taehee Lee} \\
	Department of Statistics\\
	Harvard University\\
	Cambridge, MA 02138 \\
	\texttt{taehee\_lee@fas.harvard.edu} \\
    \And
	\href{https://orcid.org/0000-0002-4450-7239}{\includegraphics[scale=0.06]{orcid.pdf}\hspace{1mm}Jun S. Liu} \\
	Department of Statistics\\
	Harvard University\\
	Cambridge, MA 02138 \\
	\texttt{jliu@stat.harvard.edu} \\
}

\renewcommand{\shorttitle}{Multi-Response Heteroscedastic Gaussian Process Models and Their Inference}

\newcommand{\KLD}[2]{\mathbb{D}_{\mathrm{KL}} \left( #1 \middle|\middle| #2 \right) }
\newcommand{\bx}{\mathbf{x}}
\newcommand{\by}{\mathbf{y}}
\newcommand{\bv}{\mathbf{v}}
\newcommand{\bz}{\mathbf{z}}
\newcommand{\bff}{\boldsymbol{f}}
\newcommand{\bg}{\boldsymbol{g}}
\newcommand{\bbf}{\boldsymbol{f}}
\newcommand{\bL}{\boldsymbol{L}}
\newcommand{\bP}{\boldsymbol{P}}
\newcommand{\rX}{X}

\newcommand{\cN}{\mathcal{N}}

\newcommand{\cGP}{\mathcal{GP}}
\newcommand{\balpha}{\boldsymbol{\alpha}}
\newcommand{\blambda}{\boldsymbol{\lambda}}
\newcommand{\bphi}{\boldsymbol{\phi}}
\newcommand{\bmu}{\boldsymbol{\mu}}
\newcommand{\vvec}{\text{vec}}
\newcommand{\diag}{\text{diag}}
\newcommand{\bbK}{\mathbb{K}}
\newcommand{\bbV}{\mathbb{V}}

\DeclareMathOperator*{\argmax}{argmax}
\DeclareMathOperator*{\argmin}{argmin}

\hypersetup{
pdftitle={Multi-Response Heteroscedastic Gaussian Process Models and Their Inference},
pdfsubject={stat.ML},
pdfauthor={Taehee Lee, Jun S. Liu},
pdfkeywords={Gaussian process, Regression, Classification, State-space models, Nonparametric heteroscedastic modelling},
}

\begin{document}
\maketitle

\begin{abstract}
Despite the widespread utilization of Gaussian process models for versatile nonparametric modeling, they exhibit limitations in effectively capturing abrupt changes in function smoothness and accommodating relationships with heteroscedastic errors. Addressing these shortcomings, the heteroscedastic Gaussian process (HeGP) regression seeks to introduce flexibility by acknowledging the variability of residual variances across covariates in the regression model. In this work, we extend the HeGP concept, expanding its scope beyond regression tasks to encompass classification and state-space models. 
To achieve this, we propose a novel framework where the Gaussian process is coupled with a covariate-induced precision matrix process, adopting a mixture formulation. This  approach enables the modeling of heteroscedastic covariance functions across covariates. To mitigate the computational challenges posed by sampling, we employ variational inference to approximate the posterior and facilitate posterior predictive modeling. Additionally, our training process leverages an EM algorithm featuring closed-form M-step updates to efficiently evaluate the heteroscedastic covariance function. A notable feature of our model is its consistent performance on multivariate responses, accommodating various types (continuous or categorical) seamlessly. Through a combination of simulations and real-world applications in climatology, we illustrate the model's prowess and advantages. By overcoming the limitations of traditional Gaussian process models, our proposed framework offers a robust and versatile tool for a wide array of applications.
\end{abstract}

\keywords{Gaussian process \and Regression \and Classification \and State-space models \and Nonparametric heteroscedastic modelling}

\section{Introduction}\label{sec1}

The Gaussian process (GP) is a flexible nonparametric modeling tool for continuous functions \citep{Rasmussen2006}. In regression analysis, the function that characterizes the relationship between the continuous covariates and responses can be modelled by a GP and the observed responses are noisy observations of the function values. In binary classification, the log-odds or probit function that defines the category probabilities can be modelled by a GP. Moreover, the hidden variables that are associated with pairs of covariates and responses in the state-space model can also be modelled by a GPs  \citep{Frigola2014,Eleftheriadis2017}.
We here focus on  multivariate GPs where   response $y$ is a $Q$-dimensional vector (it is closely related to the  multi-task Gaussian process  \citep{Bonilla2008,Leroy2020}).
Given that its attributes predominantly hinge on a singular kernel covariance function,   a GP's potential for modeling can be restrictive. In particular, one may obtain misleading results when the target function  errors display heterogeneity across covariates. 


\begin{figure}[!hbt]
\renewcommand{\baselinestretch}{0.95}
\centering
\includegraphics[width=0.41\textwidth]{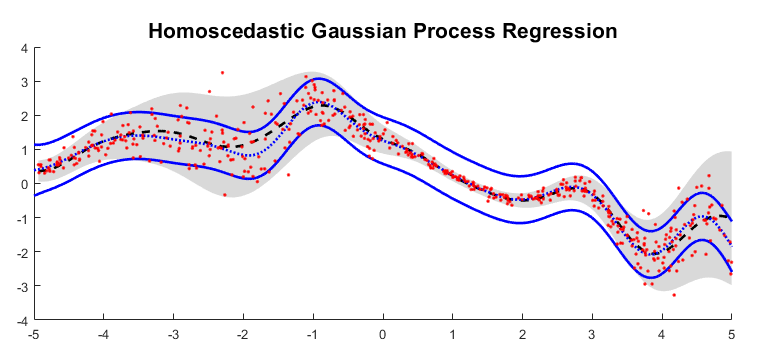}
\includegraphics[width=0.41\textwidth]{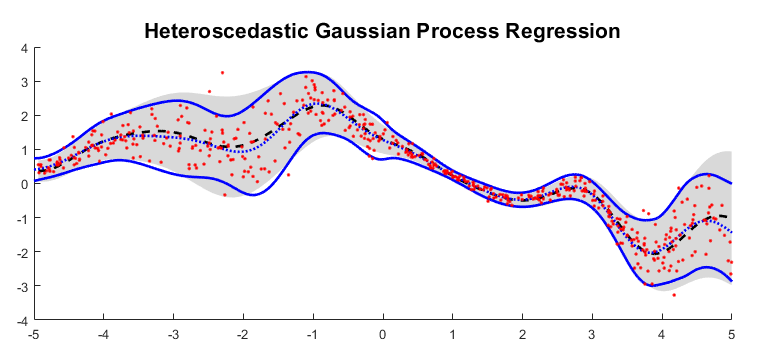}
\caption{A comparison between the HoGP and HeGP regression models on the simulated data. In each panel, shaded regions indicate the true 95\% confidence regions of the distribution that generate the data (red dots). The left and right panels show the inferred 95\% confidence bands (blue curves) from the HoGP and HeGP models, respectively.}
\label{fig_1_1}
\end{figure}

Confined to the multivariate GP regression model with Gaussian errors, the heterogeneity in the observed responses can be addressed by allowing the covariance matrix of residuals to vary over covariates, which is termed as  a heteroscedastic GP  (HeGP, henceforth) model. Figure \ref{fig_1_1} compares fittings of  the homoscedastic GP (HoGP, henceforth) and HeGP regression models on a simulated heteroscedastic time series. A main challenge in fitting a  HeGP model is  to specify how the error covariance matrix varies over covariates. 

 \cite{Wang2012}
augment the observed covariate $x$ by a latent variable $w$ and assume a GP on $(x,w)$.
For the univariate response case (i.e., $Q=1$), \cite{Le2005} model the inverse of noise variance by a GP with a positive mean and restrict it to be nonnegative;
\cite{Goldberg1998} model the logarithm of the noise variance by a GP and conducts a Bayesian analysis via Markov chain Monte Carlo (MCMC); and \cite{Lazaro-Gredilla2011} instead approximate their posterior distributions by a variational method. \cite{Kersting2007} take a similar approach as that of \cite{Goldberg1998} but use a different computational strategy via most likely noise imputation. \cite{Zhang2020} further improve the model by replacing the empirical estimate of the noise variance with an approximately unbiased estimate based on a moment estimate of regression residuals. \cite{Robani2021}  augments the k-nearest neighbor regression to overcome the overfitting issue in estimating the noise variance function. 
\cite{Binois2018} design a computationally  efficient algorithm based on \cite{Kersting2007} with a latent variable for smoothing the noise variance function.
\cite{Wang2016} follow the construction of \cite{Kersting2007} except that they construct the
heteroscedastic noise variance based on the Nadaraya-Watson kernel estimation \citep{Bierens1994,LANGRENE2019}. \cite{Lee2019} further improve the method by replacing the empirical regression of the noise variance with a closed-form function,
together with an outlier classification step.

Although the aforementioned models are important for analyzing heteroscedasticity, they have some of the following drawbacks:  1)  since some methods model the logarithm of the noise variance by
  a GP and estimate the variance by exponentiating a regression fit, 
they tend to underestimate the variability;
2)   estimation methods do not have a closed-form solution, which makes  computation very challenging;
  3) because of 
  the lack of a proper regularization, some  methods tend to overfit;
  4)  variance estimation is often intertwined with an iterative estimation procedure, which both increases computational complexity and reduces stability and validity.
Furthermore, the aforementioned models only consider the regression analysis (1-dim response), which is  limited in many applications. 

By extending an idea of \cite{Lee2019}, we generalize HeGP regression by incorporating two layers of hidden
functions: the {\it target hidden function} $\bff(\bx)$ and the {\it underlying hidden function} $\bg(\bx)$. 
Since $\bg(\bx)$ is assumed to follow a HoGP, the target hidden function $\bff(\bx)$ uses an additional  zero-mean process  with heteroscedastic covariance to model the  residuals 
not explained by $\bg(\bx)$. The response $\by$ is related to $\bff(\bx)$ through an explicit probabilistic model.
For example, a regression model assumes that   $\by$ is a linear function of  $\bff(\bx)$ plus an independent Gaussian or t-distributed error;
and a classification model uses $\bff(\bx)$ to represent the log-odds of the class-probability. 
A main contribution of this work is our treatment of the precision process, which is inspired by the idea of moving average and kernel density  estimation.

The rest of the article is organized as follows.
Section \ref{sec2} introduces our generalized HeGP model, including the model for the prior and structure for the heteroscedastic covariance process. Section~\ref{sec3} discusses the main inferential procedure and computational strategies. Section~\ref{sec4} provides more details on the application our HeGP model to robust regression, classification, and state-space modeling.  Section~\ref{sec4_0} numerically examines our method through simulations, and Section~\ref{sec5} applies our method to two real datasets. Section \ref{sec6} concludes the article with a short discussion.

\section{A General Heteroscedastic GP Model}\label{sec2}

\subsection{Common notations used throughout}
\label{sec2_1}
\begin{itemize}
\item For a matrix $A$, $vec\left(A\right)$ denotes the vectorization of $A$; for a vector $\bv$, $\text{diag}\left(\bv\right)$ denotes the diagonal matrix with diagonal entries being $\bv$;
\item $\mathcal{J}_N^{nn}$ denotes an $N\times N$ matrix of all $0$’s but a $1$ at the $\left(n,n\right)$-th entry;
and $\mathcal{J}_N^n$ denotes an $1\times N$ matrix of all $0$’s but a $1$ at the $n$th entry;
\item ${X}=\left\{\bx_n\right\}_{n=1}^{N}$ is the set of covariates, where each $\bx_n\in\mathbb{R}^{P}$ for ${P}\in\mathbb{N}$;
and ${Y}=\left\{\by_n\right\}_{n=1}^{N}$ is the set of responses with each $\by_n$ being associated with a covariate $\bx_n$.  We assume that $\by_n$ may be of multidimensional mix-type, i.e., $\by_n\in \mathcal{S}\subset \mathbb{R}^{Q}$, in which each coordinate of $\by_n$ can be  in either a finite-discrete or continuous set;
\item $\underline{X}=\left\{{\underline{\bx}}_d\right\}_{d=1}^D$ is a small set of induced covariates (tunable by the user); 
$\bL=\left\{\blambda_d\right\}_{d=1}^{D}$ is a corresponding set of $D$ unknown ${Q}\times {Q}$ positive semi-definite matrices, and $\mathcal{H}=\left\{h_d\right\}_{d=1}^{D}$, the set of unknown kernel bandwidths; we use  kernel density function $\mathcal{K}_{h}:\mathbb{R}^{P}\times\mathbb{R}^{P}\rightarrow\mathbb{R}$  with a bandwidth $h$ used to link these matrices; 
\item $\bff, \bg: \mathbb{R}^P\rightarrow\mathbb{R}^Q$, are $Q$-dimensional target  and underlying hidden functions, respectively; and
$\bmu:\mathbb{R}^P\rightarrow\mathbb{R}^Q$ denotes a $Q$-dimensional mean function;
\item For any $F$: $\mathbb{R}^{P} \rightarrow \mathbb{R}^Q$ and $X=\{\bx_n\}_{n=1}^N$, we let  $F_X$ denote the $N\times Q$ matrix with each row being $F\left(\bx_n\right)^\top$ for $\bx_n\in X$. Thus, in the same token,  $\bff_X$ and $\bg_X$ denote the $N\times Q$ matrices with their $n$th row being $\bff\left(\bx_n\right)^\top$ and $\bg(\bx_n)^\top $, respectively. 

\item Let $\bbV: \mathbb{R}^{P} \times \mathbb{R}^{P} \rightarrow \mathbb{R}^{Q \times Q}$ 
denote a multivariate kernel function. That is, for any real $N \times Q$ matrix $W$ and covariates set $X=\{\bx_n\}_{n=1}^N$ such that  $\bx_n \in \mathbb{R}^{P}$, we have ${vec\left(W\right)}^\mathbb{T}\mathbb{V}_{XX}vec\left(W\right) \geq 0$ 
holds, where $\bbV_{AB}$ denotes an $(\left|A\right|Q) \times (\left|B\right|Q)$  matrix with its 
$\left[\left(p-1\right)\left|A\right|+m,
\left(q-1\right) \left|B\right|+n\right]$-th entry being the $\left(p,q\right)$-th entry of $\mathbb{V}\left(\textbf{a}_m,\textbf{b}_n\right)$ for finite subsets $A=\left\{\textbf{a}_m\right\}_{m=1}^{\left|A\right|}$ and $B=\left\{\textbf{b}_n\right\}_{n=1}^{\left|B\right|}$ of $\mathbb{R}^{P}$. For example, $\bbV = \Sigma \otimes \bbK$ for a $Q \times Q$ positive semi-definite matrix $\Sigma$ and a kernel function $\bbK: \mathbb{R}^{P} \times \mathbb{R}^{P} \rightarrow \mathbb{R}$ satisfies the above condition, and $\bbV_{AB} = \Sigma \otimes \bbK_{AB}$  for any finite subsets $A$ and $B$;

\end{itemize}

\subsection{Description of the heteroscedastic Gaussian process model}\label{sec2_2}

Our model consists of three levels and is of the general form:
\begin{align}
\bg & \sim \cGP(\bmu,\bbV ),
\label{equation_2_0}  \\
\bbf\left(\bx\right) & \sim \mathcal{N}\left(\bg\left(\bx\right),\Lambda\left(\bx\right)\right), 
\label{equation_2_1_1}\\
\by & \sim p\left(\cdot\mid  \bbf(\bx) ; \Theta \right),
\label{equation_2_2_0} 
\end{align}
where $\Lambda:\mathbb{R}^{P}\rightarrow\mathbb{R}^{Q\times Q}$ is defined by the following inverse of the mixture of precision:
\begin{equation}
\Lambda\left(\bx\right)\triangleq\left(\sum_{d=1}^{D}{\omega_{\bx d}\blambda_d^{-1}}\right)^{-1},
\label{equation_2_3_4}
\end{equation}
and $\cGP(\ ) $ denotes a multivariate Gaussian process for a multivariate mean function $\bmu$ and kernel $\bbV$. That is, for any finite subset $Z\subseteq\mathbb{R}^{P}$, we have $\vvec(\bg_Z) \sim \cN(\vvec(\bmu_Z), \bbV_{ZZ})$. Moreover, $\omega_{\bx}$ is a $D \times 1$ weight vector with its $d$th entry  defined as:
\begin{equation}
\omega_{\bx d}\triangleq\frac{\mathcal{K}_{h_d}\left(\bx,{\underline{\bx}}_d\right)}{\sum_{c=1}^{D}{\mathcal{K}_{h_c}\left(\bx,{\underline{\bx}}_c\right)}} \propto\mathcal{K}_{h_d}\left(\bx,{\underline{\bx}}_d\right),
 \label{equation_2_1_2}
\end{equation}
which relies on a kernel $\mathcal{K}$, a set of bandwidths $\mathcal{H}$, and a set of ``induced feature vectors'' $X = \left\{{\underline{\bx}}_d\right\}_{d=1}^{D}$.
For example, when $D=1$, the distribution of $\bff (\bx)$ 
is simply multivariate Gaussian with the identical covariance matrix over covariates and the resulting model is a HoGP. When $D=2$, the distribution of $\bff(\bx)$ becomes a multivariate Gaussian distribution with a covariance matrix being a mixture of two precision matrices, $\blambda_1^{-1}$ and $\blambda_2^{-1}$, with weights proportional to  kernel distance, $\mathcal{K}_{h_i}(\bx,\underline{\bx}_i)$, $i=1,2$.
It is possible to further adjust $D$ and   $\underline{X}$
so as to fit the data better,
which is an interesting and important question in by itself but is beyond the scope of this paper. In Sections \ref{sec4} and \ref{sec5}, we fix $D=100$ in both our simulations and real data analysis, and prescribe an evenly spaced $\underline{X}$ {\it a priori} and let it fixed. Both $\bL$ and $\mathcal{H}$ will be estimated from the data. 

We call \eqref{equation_2_0} the {\it first-level} model and \eqref{equation_2_1_1} the {\it second-level} model. For an arbitrary finite subset $Z=\left\{\bz_m\right\}_{m=1}^{M}\subseteq\mathbb{R}^{P}$, we have $\vvec\left(\bbf_Z\right) \sim \mathcal{N}\left(\vvec\left(\bmu_Z\right), \mathbb{V}_{ZZ}+\Lambda_{ZZ}\right)$, where $\Lambda_{ZZ}\triangleq\sum_{m=1}^{M}\left(\Lambda\left(\bz_m\right)\otimes\mathcal{J}_M^{mm}\right)$. 
Thus, marginally $\bbf \sim \mathcal{GP} \left(\bmu, \mathbb{V}+\Lambda\right)$, which is a Gaussian process. Moreover, from  \eqref{equation_2_0} and \eqref{equation_2_1_1}, it is easy to see that  $p(\bg_X\mid \bff_X )$ has a closed-form Gaussian density.
The {\it third-level} model \eqref{equation_2_2_0} links parameters and the latent structure to the observations, which can be reflected by the conditional likelihood:
\begin{equation*}
p\left({Y}\mid \bff_X ; \Theta \right)=\prod_{n=1}^{N}{p(\by_n\mid \bff(\bx_n) ; \Theta )}.
\label{equation_2_2}
\end{equation*}
Distribution $p\left(\by_n\middle| f\left(\bx_n \right)\right)$ 
can be quite flexible. For example, where $Q=1$, we may consider a linear model with $t$-distributed errors to model outliers, e.g.,
$y_n=a\cdot f(\bx_n)+b+\epsilon_n$ for $\epsilon_n \sim  \mathcal{T}(\nu). $
We can also consider a nonlinear calibration model, e.g.,
$y_n=a\cdot \exp\left(f(\bx_n)\right)+c+\epsilon_n$ for $\epsilon_n \sim \mathcal{N}(0,1)$, 
or a binary classification model, e.g.,
$y_n \sim {\rm{Bernoulli}}\left(\frac{1}{1+\exp\left(-a\cdot f(\bx_n)-b\right)}\right).$
To make the parameters identifiable, we may need to impose certain constraint. 

\subsection{A prior on the unknown covariance matrices}\label{sec2_3}

Because \eqref{equation_2_3_4} is not identifiable and $\log{\left|\Lambda(\bx)\right|}$ which appears in the log-likelihood of the model \eqref{equation_2_1_1} is not easy to deal with in the estimation of $\blambda_d$'s, we consider the following conditional prior $p(\ \cdot\mid X)$ on $\bL$:
\begin{equation}
p\left(\bL\middle|X\right)\propto\exp\left(\frac{1}{2}\sum_{n=1}^{N}\sum_{d=1}^{D}{\omega_{\bx_nd}\log{\left|\blambda_d^{-1}\right|}}-\frac{1}{2}\sum_{n=1}^{N}\log{\left|\sum_{d=1}^{D}{\omega_{\bx_nd}\blambda_d^{-1}}\right|}\right)\cdot \pi\left(\bL\right),
\label{equation_2_3_1}
\end{equation}
where $\pi\left(\bL\right)$ is a probability density on $\bL$, such as an inverse-Wishart. Since the log-determinant function is concave, by the Jensen’s inequality we have
\begin{equation*}
\frac{1}{2}\sum_{n=1}^{N}\sum_{d=1}^{D}{\omega_{\bx_nd}\log{\left|\blambda_d^{-1}\right|}}-\frac{1}{2}\sum_{n=1}^{N}\log{\left|\sum_{d=1}^{D}{\omega_{\bx_nd}\blambda_d^{-1}}\right|}\le 0,
\label{equation_2_3_2}
\end{equation*}
which leads to, if $\pi$ is proper,
\begin{equation*}
\begin{aligned}
0 < & \int{\exp\left(\frac{1}{2}\sum_{n=1}^{N}\sum_{d=1}^{D}{\omega_{\bx_nd}\log{\left|\blambda_d^{-1}\right|}}-\frac{1}{2}\sum_{n=1}^{N}\log{\left|\sum_{d=1}^{D}{\omega_{\bx_nd}\blambda_d^{-1}}\right|}\right)\cdot\pi\left(d\bL\right)}  \le 1.
\end{aligned}
\label{equation_2_3_3}
\end{equation*}
Thus, the distribution $p(\bL \mid X)$ defined in \eqref{equation_2_3_1} is also proper.
Furthermore, $p(\bL \mid X )$ 
is maximized at
\begin{equation}
\blambda_d=\frac{\sum_{n=1}^{N}{\omega_{\bx_nd}\Lambda\left(\bx_n\right)}}{\sum_{n=1}^{N}\omega_{\bx_nd}}, \quad d=1,2,\cdots,D,
\label{equation_2_3_5}
\end{equation}
i.e., when each $\blambda_d$ is a weighted average of $\Lambda\left(x_n\right)=\left(\sum_{c=1}^{D}{\omega_{x_nc}\blambda_c^{-1}}\right)^{-1}$. Note that both \eqref{equation_2_3_4} and \eqref{equation_2_3_5} hold if all $\blambda_d$'s are identical, so this prior implicitly ``stabilizes'' the estimation of $\bL$. Moreover, 
the form of \eqref{equation_2_3_1} 
 makes the estimation of the $\blambda_d$'s easy.

\subsection{A recap}\label{sec2_4}

We consider modeling the multi-response data of the form $\left(X,Y\right)=\left\{\bx_n,\by_n\right\}_{n=1}^{N}$; where $\bx_n\in\mathbb{R}^P$ and $\by\in\mathbb{R}^Q$, using a Gaussian process.
To reflect the heteroscedasticity of the model, we have designed a three-level model. In the model, we have two multidimensional hidden functions, $\bg$ and $\bff$. While $\bg$ follows a (homoscedastic) multivariate Gaussian process over the covariates $X$, $\bff$ is linked to the corresponding observations $Y$ through a parametric model such as logistic. The heteroscedasticity of the model is defined by the relationship between $\bg$ and $\bff$. The model \eqref{equation_2_1_1} assumes that $\bff$ given $\bg$ follows a Gaussian distribution with $\bg$ as its mean and $\Lambda^{-1}$, which is a mixture of precision matrices $\blambda_{d}^{-1}$, as its precision matrix. The weights of mixtures $\omega_{\bx} = \left\{\omega_{\bx d}\right\}_{d=1}^D$ are defined at the input covariate $\bx$ by a density kernel $\mathcal{K}$, bandwidths $\mathcal{H}$, and distances from that covariate $\bx$ to a set of given fixed induced covariates $\underline{X} = \left\{{\underline{\bx}}_d\right\}_{d=1}^D$.
Under this model, the unknown parameters to be estimated are kernel hyperparameters of $\bbV$ (for example, if $\bbV = \Sigma \otimes \bbK$ then $\Sigma$ and $\bbK$ are to be estimated), parameters for the mean function $\bmu(\bx)$, parameters in the third level connecting $\by_n$ to $f$, and the set of covariance matrices $\bL=\left\{\blambda_d\right\}_{d=1}^D$. The kernel function $\mathcal{K}$ is assumed known but the set of bandwidths $\mathcal{H}$ for defining $\omega_{\bx}$ are to be estimated.

The above model has both  theoretical and  practical challenges. Because our model does not directly depend on  the $\blambda_d$'s but rather their mixture, $\Lambda$, the set $\bL$ might not be identifiable, especially if $D$ is larger than the size of the training data $N$. Furthermore, the term $\log{\left|\Lambda\left(\bx_n\right)\right|} = \log{\left|\sum_{d=1}^{D}{\omega_{\bx_n d}\blambda_d^{-1}}\right|}$, which appears in the log-likelihood of  model \eqref{equation_2_1_1}, is not easy to handle. 
Also, some of the $\blambda_d$'s may be estimated as singular matrices during the estimation step, which makes the inference unstable. To address these challenges, we design a prior distribution \eqref{equation_2_3_1} on $\bL$, conditioning on the training covariates $X$, which forces $\blambda_d$'s to be ``closer'' to each other and converts  $\log{\left|\sum_{d=1}^{D}{\omega_{\bx_n d}\blambda_d^{-1}}\right|}$ into $\sum_{d=1}^{D}{\omega_{\bx_{n}d}\log{\left|\blambda_d^{-1}\right|}}$ in the log-likelihood.

For the three-level heteroscedastic Gaussian process model in Section \ref{sec2_2} endowed with the prior in Section \ref{sec2_3}, we show in Section \ref{sec3} how to construct a variational EM (VEM) algorithm to estimate $\bL$ and other unknown parameters.
A particularly attractive feature of our VEM approach is that the M-step update of $\bL$ is given in a \textit{closed}-form for some special choices of the prior $\pi$ in \eqref{equation_2_3_1}, such as a flat  or an inverse-Wishart distribution.

\section{Inference of the Model Parameters}\label{sec3}

A main challenge in our inference framework is to estimate (or determine) the following four categories of parameters: 1) $\bL$, the set of covariance matrices, 2) $\mathcal{H}$, the set of density kernel bandwidths, 3) $\Theta$, the set of parameters of the third-level model in \eqref{equation_2_2_0}, and 4) $\Upsilon$, the set of kernel hyperparameters of $\bbV$ and parameters of $\mu$ in \eqref{equation_2_0}. 

If the set of kernel bandwidths $\mathcal{H}$ is given {\it a priori}, we can follow a variational EM (VEM) strategy to estimate $\bL$, $\Theta$ and $\Upsilon$. However, in practice, $\mathcal{H}$ is usually unknown as well as the others, so our algorithm includes its estimation step as well. The E-step aims at approximating the posterior by optimizing an evidence lower bound (ELBO) with respect to the variational parameters and the M-step is for updating $\Theta$ and $\Upsilon$. $\mathcal{H}$ is determined by a cross-validation after each iteration of the EM algorithm.

In Section \ref{sec3_1} we describe a variational EM algorithm for estimating $\bL$, $\Theta$ and $\Upsilon$, and discuss the prediction issue in Section \ref{sec3_6}, with details that can be found in Appendix: Appendix \ref{secA_1} and \ref{secA_2} detail its E-step and M-step, respectively, and Appendix \ref{secB1} provides a procedure to determine bandwidths $\mathcal{H}=\left\{h_d\right\}_{d=1}^{D}$.

\subsection{A variational EM algorithm for parameter estimation} \label{sec3_1}

As mentioned earlier, bandwidths in  $\mathcal{H}$  are assumed given and not estimated in the following estimation procedure, so  we omit it in the formulation for convenience. The ``complete-data'' likelihood can be written as
\begin{equation}
\begin{aligned}
p \left({Y},\bbf_{X},\bg_{X},\bL \middle| X;\Theta,\Upsilon\right) = p\left({Y}\middle| \bff_X;\Theta\right) p\left(
\bff_X\middle| \bg_X;\bL\right) p\left(\bg_X\middle|\Upsilon\right) p\left(\bL \middle| X \right),
\end{aligned}
\label{equation_3_1_1}
\end{equation}
where the distributions in \eqref{equation_3_1_1} are explained and defined in \eqref{equation_2_2_0}, \eqref{equation_2_1_1}, \eqref{equation_2_0}, and \eqref{equation_2_3_1}, respectively in order.
For the third-level model, we have $p\left({Y}\middle| \bbf_X;\Theta\right) \triangleq\prod_{n=1}^{N}{p\left(\by_n\middle| \bbf\left(\bx_n\right);\Theta\right)}.$
The latent structures have the distributions:
\begin{equation*}
\begin{aligned}
p\left(\bbf_X\mid \bg_X;L\right)\triangleq \prod_{n=1}^{N}{\mathcal{N}\left(\bbf\left(\bx_n\right)\middle| \bg\left(\bx_n\right),\Lambda\left(\bx_n\right)\right)} \ \ \mbox{and} \ \
p\left(\bg_X\middle| \Upsilon \right)  \triangleq   \mathcal{N}\left(\vvec\left(\bg_X\right) \middle| \vvec\left(\bmu_{X}\right), \bbV_{XX}\right).
\end{aligned}
\label{equation_3_1_2}
\end{equation*}

A typical approach for the estimation is to maximize the {\it observed-data} likelihood, which requires one to integrate out both $f_X$ and $g_X$. However, if the third level model $p\left(Y\middle|\bff_X,\Theta\right)$ is not Gaussian, then the observed-data (or marginal) likelihood $p\left({Y}\middle|\bL,\Theta \right)$ might not be expressed in a  closed-form, making its optimization with respect to $\Theta$ intractable. We thus consider augmenting the hidden variables $\bff_X$ and $\bg_X$ and implementing an EM-type of approach. The 
standard EM algorithm should proceed as follows:


\begin{itemize}
\item E-step - at iteration $t$, we compute:
\begin{equation}
\begin{aligned}
\mathcal{Q}\left(\bL,\Theta,\Upsilon\middle|\bL^{\left(t\right)},{\Theta}^{\left(t\right)},{\Upsilon}^{\left(t\right)}\right) \triangleq\mathbb{E}_{p\left(\bff_X,\bg_X\mid {Y};\bL^{\left(t\right)},{\Theta}^{\left(t\right)},{\Upsilon}^{\left(t\right)}\right)}\left[\log{p\left({Y},\bff_X,\bg_X,\bL \middle| X ; {\Theta},\Upsilon \right)}\right].
\end{aligned}
\label{equation_3_1_3}
\end{equation}
\item M-step - we update 
$\bL$, $\Theta$ and $\Upsilon$ as 
\begin{equation*}
\left(\bL^{\left(t+1\right)},{\Theta}^{\left(t+1\right)},{\Upsilon}^{\left(t+1\right)}\right)\triangleq\argmax_{\bL,\Theta,\Upsilon}{\mathcal{Q}\left(\bL,\Theta,\Upsilon \middle| \bL^{\left(t\right)},{\Theta}^{\left(t\right)},{\Upsilon}^{\left(t\right)}\right)}.
\label{equation_3_1_4}
\end{equation*}
\end{itemize}


However, the posterior distribution 
$p\left(\bbf_X,\bg_X \ \middle| \ X, Y;\bL^{(t)},{\Theta}^{\left(t\right)},{\Upsilon}^{\left(t\right)} \right)$ may not be in a nice form if the third-level model $p\left(Y\middle| \bbf_X;\Theta^{\left(t\right)}\right)$ in \eqref{equation_2_2_0} is not Gaussian. Thus, it is not feasible to compute the exact Q-function, $\mathcal{Q}(\bL,\Theta, \Upsilon \mid\bL^{\left(t\right)},{\Theta}^{\left(t\right)},{\Upsilon}^{\left(t\right)})$, in general. To obtain an iterative algorithm with a {closed-form} update 
in the M-step, we resort to an approximation instead. To be specific, we note that:
\begin{equation}
p\left(\bbf_X,\bg_X\middle|Y,X;\bL,\Theta,\Upsilon\right)=p\left(\bbf_X\middle|Y,X;\bL,\Theta,\Upsilon\right)p\left(\bg_X\middle| \bbf_X;\bL,\Upsilon\right),
\label{equation_3_1_5}
\end{equation}
where $p\left(\bg_X\middle| \bbf_X;\bL,\Upsilon\right) \propto p\left(\bbf_X\middle| \bg_X;\bL\right)p\left(\bg_X\middle|\Upsilon\right)$ is a Gaussian distribution (in closed-form), but $p\left(\bbf_X\middle|Y,X;\bL,\Theta,\Upsilon\right)$ is often non-standard if $p\left(Y\mid \bbf_X;\Theta\right)$ is non-Gaussian. Although we may approximate the marginal likelihood by Markov chain Monte Carlo 
(see Appendix \ref{secD}), which is computationally costly, we here describe a variational approximation method \citep{Neal1998}. 
With $\Gamma = (\eta,\Psi)$ being the set of variational parameters, we consider the approximation
\begin{equation}
p\left(\bbf_X\middle|Y,X;\bL^{\left(t\right)},\Theta^{\left(t\right)},\Upsilon^{\left(t\right)}\right) \approx q\left(\bbf_X\middle|\Gamma\right) \triangleq \prod_{n=1}^{N}{\mathcal{N}\left(\bbf\left(\bx_n\right)\mid{\eta}_n,{\Psi}_n\right)},
\label{equation_3_1_6}
\end{equation}
where each $\eta_n$ and $\Psi_n$ are a $Q \times 1$ vector parameter and $Q \times Q$ covariance parameter, respectively. Our goal is to find a $ \Gamma$ 
that minimizes the  Kullback-Leibler (KL) divergence:
\begin{equation}
\hat{\Gamma}^{\left(t\right)} \triangleq  \argmin_{\Gamma}{ \KLD{q\left(\bbf_X\middle|\Gamma\right)}{p\left(\bbf_X\middle|Y,X;\bL^{\left(t\right)},\Theta^{\left(t\right)},\Upsilon^{\left(t\right)}\right)}}.
\label{equation_3_1_7}
\end{equation}
The reparameterization trick \citep{Kingma2013} may be applied when using Monte Carlo to approximate the KL divergence. 

Once $\Gamma = \hat{\Gamma}^{(t)}$ is found, the Q-function is approximated as:
\begin{equation}
\begin{aligned}
\mathcal{Q}\left(\bL,\Theta,\Upsilon\middle|\bL^{\left(t\right)},{\Theta}^{\left(t\right)},{\Upsilon}^{\left(t\right)}\right) \approx \mathbb{E}_{q\left(\bbf_X,\bg_X\middle|\Upsilon^{\left(t\right)},\bL^{\left(t\right)},{\hat{\Gamma}}^{\left(t\right)}\right)}\left[\log{p\left({Y},\bff_X,\bg_X,\bL \middle| X ; {\Theta},\Upsilon \right)}\right],
\end{aligned}
\label{equation_3_1_9}
\end{equation}
in which $q\left(\bbf_X,\bg_X\middle|\Upsilon^{\left(t\right)},\bL^{\left(t\right)},{\hat{\Gamma}}^{\left(t\right)}\right) \triangleq q\left(\bbf_X\middle|{\hat{\Gamma}}^{\left(t\right)}\right)p\left(\bg_X\middle| \bbf_X,\Upsilon^{\left(t\right)};\bL^{\left(t\right)}\right)$ is used in the place of the ``correct'' one, i.e., $p\left(\bbf_X,\bg_X\middle|Y,X;\bL^{\left(t\right)},\Theta^{\left(t\right)},\Upsilon^{\left(t\right)}\right)$.

In each M-step, we update $\Theta$ and $\Upsilon$ numerically, and $\bL$ with a closed-form update. To be specific, we rely on the following \textit{approximated} objective functions:
\begin{equation}
\begin{aligned}
{\Theta}^{\left(t+1\right)} & \triangleq \argmax_{\Theta}{ \mathbb{E}_{q\left(\bbf_X\middle|\hat{\Gamma}^{\left(t\right)}\right)}\left[\log{p\left(Y\middle| \bbf_X,\Theta\right)}\right]} \\ {\Upsilon}^{\left(t+1\right)} & \triangleq \argmax_{\Upsilon}{\mathbb{E}_{q\left(\bbf_X,\bg_X\middle|\bL^{\left(t\right)},\Upsilon^{\left(t\right)},\hat{\Gamma}^{\left(t\right)}\right)}\left[\log{p\left(\bg_X\middle|\Upsilon\right)}\right]} \\ {\bL}^{\left(t+1\right)} & \triangleq \argmax_{\bL}{\mathbb{E}_{q\left(\bbf_X,\bg_X\middle|\bL^{\left(t\right)},\Upsilon^{\left(t\right)},\hat{\Gamma}^{\left(t\right)}\right)}\left[\log{p\left(\bbf_X\middle| \bg_X ;\bL\right)} + \log{p\left(\bL\middle|X\right)} \right]},
\end{aligned}
\label{equation_3_1_11}
\end{equation}
for the estimated variational parameter $\hat{\Gamma}^{\left(t\right)}$ in the E-step. Thus, the updates of $\Theta$, $\bL$ and $\Upsilon$ are done {\it{separately}}.

In summary, although $\bL$ can be very high-dimensional, it does not lead to overfitting due to the additional ``smoothing'' step induced by the model \eqref{equation_2_3_4} and the prior \eqref{equation_2_3_1}. We resort to a variational EM algorithm \citep{Neal1998,Jordan1999} to derive parameter estimates, relying on both density  approximations and entropy lower bounds. Details of the inference can be found in Appendix \ref{secA} and \ref{secB}.

The variational EM approach we just described takes as given the bandwidths $\mathcal{H}=\left\{h_d\right\}_{d=1}^{D}$ of the density kernel $\mathcal{K}$, used  for modeling the precision matrix process. Note that the modeling is very sensitive to the choice of $\mathcal{H}$ which controls the smoothness of the heteroscedasticity. Following the nearest-neighbor bandwidth idea of \cite{LANGRENE2019}, we choose each $h_d$ as a radius for the corresponding neighborhood of $\underline{\bx}_d$ to cover a certain proportion of the input covariates $X=\left\{\bx_n\right\}_{n=1}^{N}$, and the proportion is chosen by the cross-validation. Details can be found in Appendix \ref{secB1}.

\subsection{Algorithm}\label{sec3_6}

The ``vanilla'' version of our method is summarized in Algorithm \ref{algorithm1}, and one may add more layers to the model. 
Once the maximum number of the iterations  is reached, the estimated $\hat{\Theta}$, $\hat{\Upsilon}$, $\hat{\bL}$, $\hat{\Gamma}$ and $\hat{\mathcal{H}}$ are returned, from which one can derive a closed-form approximated posterior predictive distribution of the hidden function $g\left(\bx\right)$ for an arbitrary $\bx$ as:
\begin{equation}
\begin{aligned}
& p\left(\bg\left(\bx\right)\ \middle| \  Y, \hat{\bL},\hat{\Theta},\hat{\Upsilon},\hat{\mathcal{H}}\right) = \mathbb{E}_{p\left(\bg_X \middle| Y;\hat{\bL},\hat{\Theta},\hat{\Upsilon},\hat{\mathcal{H}}\right)}\left[p\left(\bg\left(\bx\right) \ \middle| \ \bg_X;\hat{\Upsilon}\right)\right] \\ & \approx \mathbb{E}_{q\left(\bg_X\middle|\hat{\bL},\hat{\Upsilon},\hat{\mathcal{H}},\hat{\Gamma}\right)}\left[p\left(\bg\left(\bx\right) \ \middle| \ \bg_X;\hat{\Upsilon}\right)\right] = \mathcal{N}\left(\bg\left(\bx\right) \ \middle| \ \overline{\mu}\left(\bx\right),\overline{\nu}\left(\bx\right)\right),
\end{aligned}
\label{equation_3_4_0}
\end{equation}
where $p\left(g\left(\bx\right)\middle| g_X;\hat{\Upsilon}\right)$ is a conditional Gaussian process derived from \eqref{equation_2_1_1} and:
\begin{equation*}
\begin{aligned}
\overline{\mu}\left(x\right) & 
\triangleq \hat{\mathbb{V}}_{\bx X}\left(\hat{\mathbb{V}}_{XX}+\hat{\Lambda}_{XX}\right)^{-1}\left(\vvec\left({\hat{\eta}}_X\right)-\vvec\left(\hat{\mu}_X\right)\right)+\vvec\left(\hat{\mu}_X\right) \\ \overline{\nu}\left(\bx\right) & \triangleq \hat{\mathbb{V}}_{\bx \bx}-\hat{\mathbb{V}}_{\bx X}\left(\hat{\mathbb{V}}_{XX}+\hat{\Lambda}_{XX}\right)^{-1}\hat{\mathbb{V}}_{X \bx} + \hat{\mathbb{V}}_{\bx X}\left(\hat{\mathbb{V}}_{XX} + \hat{\Lambda}_{XX}\right)^{-1}{\hat{\Psi}}_{XX}\left(\hat{\mathbb{V}}_{XX}+\hat{\Lambda}_{XX}\right)^{-1}\hat{\mathbb{V}}_{X \bx},
\end{aligned}
\label{equation_3_4_2}
\end{equation*}
where $\hat{\Lambda}$ is defined as \eqref{equation_2_3_4}, from the estimated $\hat{\bL}$.

From \eqref{equation_3_4_0}, the posterior predictive distribution of the hidden function $\bbf\left(\bx\right)$ for an arbitrary $\bx$ can be approximated by a Gaussian distribution in \textit{closed}-form as follows:
\begin{equation}
\begin{aligned}
& p\left(\bbf\left(\bx\right)\middle|Y;\hat{L},\hat{\Theta},\hat{\Upsilon},\hat{\mathcal{H}}\right) = \mathbb{E}_{p\left(\bg\left(\bx\right)\middle|Y;\hat{\bL},\hat{\Theta},\hat{\Upsilon},\hat{\mathcal{H}}\right)}\left[p\left(\bbf\left(\bx\right)\middle| \bg\left(\bx\right);\hat{\bL},\hat{\mathcal{H}}\right)\right] \\ & \approx \mathbb{E}_{\mathcal{N}\left(\bg\left(\bx\right)\middle|\overline{\mu}\left(\bx\right),\overline{\nu}\left(\bx\right)\right)}\left[p\left(\bbf\left(\bx\right)\middle| \bg\left(\bx\right);\hat{\bL},\hat{\mathcal{H}}\right)\right] = \mathcal{N}\left(\bbf\left(\bx\right)\middle|\overline{\mu}\left(\bx\right),\overline{\nu}\left(\bx\right)+{\hat{\Lambda}\left(\bx\right)}\right),
\end{aligned}
\label{equation_3_4_1}
\end{equation}
where $\hat{\Lambda}\left(\bx\right) \triangleq \left(\sum_{d=1}^{D}{{\hat{\omega}}_{\bx d}{\hat{\blambda}}_d^{-1}}\right)^{-1}$.
From \eqref{equation_3_4_1}, we have an approximated posterior predictive distribution of $\by$ at $\bx$:
\begin{equation*}
\begin{aligned}
& p\left(\by\middle|Y;\hat{\bL},\hat{\Theta},\hat{\Upsilon},\hat{\mathcal{H}}\right) = \int{p\left(\by\middle| \bbf\left(\bx\right);\hat{\Theta}\right)p\left(\bbf\left(\bx\right)\middle|Y;\hat{\bL},\hat{\Theta},\hat{\Upsilon},\hat{\mathcal{H}}\right){d\bbf\left(\bx\right)}} \\ & \approx \int{p\left(\by\middle| \bbf\left(\bx\right);\hat{\Theta}\right)\mathcal{N}\left(\bbf\left(\bx\right)\middle|\overline{\mu}\left(\bx\right),\overline{\nu}\left(\bx\right)+{\hat{\Lambda}\left(\bx\right)}\right){d\bbf\left(\bx\right)}}.
\end{aligned}
\label{equation_3_4_3}
\end{equation*}

More general cases are discussed in the Appendix. For example, Appendix \ref{secC}  describes  a Bayesian estimation strategy;  Appendix \ref{secE} investigates the case where some responses contain missing values; and  Appendix \ref{secF} generalizes our model to handle heterogeneous response components. Additional variational approximations for dealing with large training datasets are discussed in Appendix \ref{secG}. These extensions can be reflected in constructing a ``general'' version of the algorithm for applications. To verify that our algorithm is both practical and reasonably accurate, we apply its special cases discussed in Section \ref{sec4} to both simulated and real datasets as shown in Sections \ref{sec4_0} and \ref{sec5}.

\begin{algorithm}[!hbt]
\SetAlgoLined
\DontPrintSemicolon
 \textbf{initialize} the set of parameters $\Theta=\Theta^{\left(0\right)}$, the set of variational parameters $\Gamma$, the set of kernel hyperparameters $\Upsilon=\Upsilon^{\left(0\right)}$, bandwidth percentage $r = \hat{r}$ in the set of candidates $\mathcal{R}$, and $\bL=\bL^{\left(0\right)}$ as a set of identical constant matrices.\;
 \While{convergence}{
  \textbf{(Bandwidth) update} bandwidths $\mathcal{H}=\hat{\mathcal{H}}$ for $r = \hat{r}$ by the 'K-nearest-neighbors' rule in Appendix \ref{secB1}.\;
  \textbf{(E-step) estimate} $\Gamma=\hat{\Gamma}$ by gradient ascent on the objective function ELBO in \eqref{equation_3_1_7}, given $\bL=\bL^{\left(t\right)}$, $\Theta=\Theta^{\left(t\right)}$, $\Upsilon=\Upsilon^{\left(t\right)}$, and $\mathcal{H}=\hat{\mathcal{H}}$.\;
  \textbf{(M-step) update} $\bL=\bL^{\left(t+1\right)}$ by \eqref{equation_A_2_11} and $\Theta=\Theta^{\left(t+1\right)}$ and $\Upsilon=\Upsilon^{\left(t+1\right)}$ by gradient ascent on the objective functions in \eqref{equation_A_2_2}, given $\Gamma=\hat{\Gamma}$ and $\mathcal{H}=\hat{\mathcal{H}}$.\;
  \textbf{(Cross-validation) update} $r = \hat{r} \in \mathcal{R}$ by the cross-validation discussed in Appendix \ref{secB1}, given the estimated $\bL=\bL^{\left(t+1\right)}$, $\Theta=\Theta^{\left(t+1\right)}$, $\Upsilon=\Upsilon^{\left(t+1\right)}$, and $\Gamma=\hat{\Gamma}$.\;
 }
 \caption{Parameter Estimation for Generalized HeGP}
\label{algorithm1}\end{algorithm}



\section{Regression, Classification, and State-Space Models}\label{sec4}



\subsection{Regression modeling}\label{sec4_1}

Gaussian process regression (GPR) is popular because of its flexibility in fitting nonlinear relationships and its elegant closed-form solution. As mentioned in Section \ref{sec1}, although the error distribution in this  model is Gaussian, its variance (or covariance matrix for multidimensional responses) may vary along with the covariates. Also, in practice the responses may be contaminated by outliers, or the residuals 
may follow a thick-tailed distribution such as a Student's t-distribution. In this section, we discuss how our HeGP modeling strategy can be applied to accommodate these complications.

\subsubsection{Gaussian residuals and HeGPR-G method}\label{sec4_1_1}

A typical GP regression model with Gaussian errors can be expressed as a compressed form of \eqref{equation_2_2_0} by simply letting
 $\by\equiv \bbf\left(\bx\right)$. We call the corresponding method  ``HeGPR-G."  In this way, the three-layer model in Section \ref{sec2_2} is reduced to the following  two-layer one:
\begin{align*}
\bg  \sim \cGP\left(\bmu,\bbV \right)   \ \ \mbox{and} \ \   
\by \sim \mathcal{N}\left(\bg\left(\bx\right),\Lambda\left(\bx\right)\right).
\end{align*}
The  ``complete-data'' likelihood becomes
\begin{equation*}
p\left(Y,\bg_X,\bL \middle| X; \Upsilon\right) = p\left(Y\middle| \bg_X;\bL\right) p\left(\bg_X\middle|\Upsilon\right) p\left(\bL \middle| X \right),
\label{equation_4_0_5}
\end{equation*}
where $p \left(Y\middle| \bg_X; \bL\right)$ 
is a closed-form Gaussian distribution. The Q-function is
\begin{equation*}
\mathcal{Q}\left(\bL,\Upsilon\middle|\bL^{\left(t\right)},\Upsilon^{\left(t\right)}\right) = \mathbb{E}_{p\left(\bg_X\middle|Y;\bL^{\left(t\right)},\Upsilon^{\left(t\right)}\right)}\left[\log{p\left(Y,\bg_X,\bL\middle|X,\Upsilon\right)}\right],
\label{equation_4_0_6}
\end{equation*}
which gives rise to the objective functions in the M-step as:
\begin{equation}
\begin{aligned}
{\Upsilon}^{\left(t+1\right)} & \triangleq \argmax_{\Upsilon}{\mathbb{E}_{p\left(\bg_X\middle|Y;\bL^{\left(t\right)},\Upsilon^{\left(t\right)}\right)}\left[\log{p\left(\bg_X\middle|\Upsilon\right)}\right]} \\ {\bL}^{\left(t+1\right)} & \triangleq \argmax_{\bL}{\mathbb{E}_{p\left(\bg_X\middle|Y;\bL^{\left(t\right)},\Upsilon^{\left(t\right)}\right)}\left[\log{p\left(Y\middle| \bg_X;\bL\right)} + p\left(\bL\middle|X\right) \right]}.
\end{aligned}
\label{equation_4_0_7}
\end{equation}

Note that \eqref{equation_4_0_7}
leads to  an exact EM algorithm. Though the update for $\Upsilon$ still requires a numerical method, one can derive the same closed-form update for $\bL^{(t+1)}$ as in \eqref{equation_A_2_11} with the same $\mathbb{A}$ and $\mathbb{B}$ in \eqref{equation_A_2_9}, but a different $\Omega_X = \vvec\left(\textbf{Y}-\hat{\mu}_{X}^{\left(t\right)}\right){\vvec\left(\textbf{Y}-\hat{\mu}_{X}^{\left(t\right)}\right)}^\top$, where $\textbf{Y}=(\by_1,\ldots, \by_N)^\top$, an $N \times Q$ matrix.  

\subsubsection{Handling thick-tailed residuals}\label{sec4_1_3}

We extend the model of Section \ref{sec4_1_1} to accommodate thick-tailed residuals as follows:
\begin{align}
\bg & \sim \cGP(\bmu,\bbV ),
\label{equation_4_1_3_1}  \\
\bbf\left(\bx\right) & \sim \mathcal{N}\left(\bg\left(\bx\right),\Lambda\left(\bx\right)\right), 
\label{equation_4_1_3_2}\\
\by & \sim \mathcal{T}_\nu\left(\bbf\left(\bx\right),\Phi\left(\bx\right)\right),
\label{equation_4_1_3_3} 
\end{align}
where $\mathcal{T}_\nu$ represents the multivariate t-distribution with $\nu$ degrees of freedom, and
$\Lambda^{-1}\left(\bx\right)\triangleq \sum_{d=1}^{D}{\omega_{\bx d}\blambda_d^{-1}}$ and $\Phi^{-1}\left(\bx\right)\triangleq
\sum_{d=1}^{D}{\omega_{\bx d}\bphi_d^{-1}}$ are constructed from unknown $Q\times Q$  base  matrices $\bL = \left\{\blambda_d\right\}_{d=1}^D$ and $\bP = \left\{\bphi_d\right\}_{d=1}^D$, respectively. The resulting method is denoted as HeGPR-H. 
Note that $\Lambda \equiv 0$ leads to  a GR regression model with Student's t-residuals, and $\Phi \equiv 0$ corresponds to HeGPR-G.

With a set of scale random scales $\balpha=\left\{\alpha_n\right\}_{n=1}^N$, we can represent t-distributed random variables as scale-mixtures of Gaussian random variables and have the following ``complete-data'' likelihood:
\begin{equation*}
\begin{aligned}
p \left({Y},\bf\bff_X,\bg_X, \bL, \balpha, \bP \middle| X;\Upsilon\right)  = p\left({Y}\middle| \bbf_X, \balpha; \bP \right) p\left(
\bbf_X\middle| \bg_X ;\bL\right) p\left(\bg_X\middle|\Upsilon\right) p\left(\balpha\right) p\left(\bL \middle|X\right) p\left(\bP \middle|X\right),
\end{aligned}
\label{equation_4_1_3_4}
\end{equation*}
where the new latent structures $p\left({Y}\middle| \bbf_X, \balpha; \bP \right)$, $p\left(\balpha\right)$ and $p\left(\bP \middle|X\right)$ have the following distributions:
\begin{equation}
\begin{aligned}
p\left({Y}\middle| \bbf_X, \balpha; \bP \right) & \triangleq \prod_{n=1}^{N}{\mathcal{N}\left(\by_{n}\middle|\bbf\left(\bx_n\right),\alpha_{n}\cdot \Phi\left(\bx_n\right)\right)} \\ 
p\left(\balpha\right) & = \prod_{n=1}^{N}p\left(\alpha_n\right) \triangleq \prod_{n=1}^{N}{\mathcal{IG}\left(\alpha_n\middle|\frac{\nu}{2},\frac{\nu}{2}\right)} \\  p\left(\bP\middle|X\right) & \propto \exp\left(\frac{1}{2}\sum_{n=1}^{N}\sum_{d=1}^{D}{\omega_{\bx_nd}\log{\left|\bphi_d^{-1}\right|}}-\frac{1}{2}\sum_{n=1}^{N}\log{\left|\sum_{d=1}^{D}{\omega_{\bx_nd}\bphi_d^{-1}}\right|}\right).
\end{aligned}
\label{equation_4_1_3_5}
\end{equation}
Next, we approximate the posterior distribution, $p (\bff_X,\bg_X, \balpha \mid Y,X,\bP^{\left(t\right)},\bL^{\left(t\right)},\Upsilon^{\left(t\right)})$, by $q(\cdot)$ to accomplish the variational E-step:
\begin{equation}
\begin{aligned}
q(\bbf_X,\bg_X,\balpha\mid\bL^{(t)},\Upsilon^{(t)}, \Gamma) \triangleq q(\bbf_X,\balpha \mid \Gamma) \cdot p(\bg_X\mid \bbf_X,\bL^{(t)},\Upsilon^{(t)}),
\end{aligned}
\label{equation_4_1_3_6}
\end{equation}
where  $q(\bff_X,\balpha \mid\Gamma)=q\left(\bff_X | \Gamma\right)q\left(\balpha | \Gamma\right)$ with
\begin{equation}
\begin{aligned}
q\left(\bbf_X\mid\Gamma\right) \triangleq \prod_{n=1}^{N}{\cN\left(\bbf\left(\bx_n\right) \middle|\eta_{n},\Psi_{n}\right)} \ \ \mbox{and} \ \
q\left(\balpha\mid\Gamma\right) \triangleq \prod_{n=1}^{N}{\mathcal{IG}\left(\alpha_n\middle|\frac{\nu + Q}{2},\frac{\nu + Q}{2} \xi_{n}^{-2}\right)}.
\end{aligned}
\label{equation_4_1_3_7}
\end{equation}
Here $\Gamma \triangleq \left\{\eta,\Psi,\xi\right\}$ is the set of variational parameters, with $\eta\triangleq\left\{\eta_{n}\right\}_{n=1}^{N}$, $\Psi\triangleq\left\{\Psi_{n}\right\}_{n=1}^{N}$ and $\xi\triangleq\left\{\xi_{n}\right\}_{n=1}^{N}$. While we have the same $q\left(\bbf_X | \Gamma\right)$ as in \eqref{equation_3_1_6}, function  $q\left(\balpha | \Gamma\right)$ in \eqref{equation_4_1_3_7} reflects the dimensionality $Q$ of the responses.  Conditioning on $\bff_X$, the exact distribution of $\alpha_n$ has this form, but has its $\xi_n$ being a nonlinear function of $\bff_X$. We thus let $\xi_n$'s be a  free variational parameters to derive a good approximation to the Q-function.

With the above variational approximation, the closed-form M-step updates for $\bP^{\left(t+1\right)}$ and $\bL^{\left(t+1\right)}$ are available as follows:
\begin{equation}
\begin{aligned}
\bphi_d^{\left(t+1\right)} & = \frac{\sum_{n=1}^{N}{\omega_{\bx_nd}\left({\hat{\xi}}_n^{\left(t\right)}\right)^2 \left(\left(\by_n-{\hat{\eta}}_n^{\left(t\right)}\right)\left(\by_n-{\hat{\eta}}_n^{\left(t\right)}\right)^\mathbb{T}+{\hat{\Psi}}_n^{\left(t\right)}\right) }}{\sum_{n=1}^{N}\omega_{\bx_nd}} \\ 
\blambda_d^{\left(t+1\right)} & =\frac{\sum_{n=1}^{N}{\omega_{\bx_nd}\left(\mathbb{A}_{\bx_n}+\mathbb{B}_{\bx_n}\Omega_X\mathbb{B}_{\bx_n}^\mathbb{T}\right)}}{\sum_{n=1}^{N}\omega_{\bx_nd}},
\end{aligned}
\label{equation_4_1_3_12}
\end{equation}
with the same $\mathbb{A}$, $\mathbb{B}$ and $\Omega$ defined in \eqref{equation_A_2_9}. 

An intuition behind the model formulation \eqref{equation_4_1_3_5} and the variational approximation \eqref{equation_4_1_3_6} and \eqref{equation_4_1_3_7} is that,  {given} $\xi_n = \hat{\xi}_n$'s and the regression function $\bg$, we can approximate the rest 
by a simple Gaussian model (see Appendix \ref{secB2} for details):
\begin{equation}
\begin{aligned}
\bbf\left(\bx_n\right) \sim \mathcal{N}\left(\bg\left(\bx_n\right),\Lambda\left(\bx_n\right)\right)  \ \ \mbox{and} \ \ \by_n \sim \mathcal{N}\left(\bbf\left(\bx_n\right),\hat{\xi}_n^{-2}\cdot\Psi\left(\bx_n\right)\right),  
\end{aligned}
\label{equation_4_1_3_15}
\end{equation}
which gives rise to  $\by_n \sim \mathcal{N}\left(\bg\left(\bx_n\right),\hat{\xi}_n^{-2}\cdot\Psi\left(\bx_n\right)+\Lambda\left(\bx_n\right)\right)$. 
As a consequence, we can 
derive the closed-form M-step updates as \eqref{equation_4_1_3_12}. 

\subsubsection{Gaussian residuals with outliers}\label{sec4_1_4}

One special case of the HeGPR-H in Section \ref{sec4_1_3} is to assume in the third-level that 
$\Phi\left(\bx\right) \equiv \sigma_{0}^{2} \cdot \Lambda\left(\bx\right)$, for a scalar hyperparameter $0 < \sigma_0^{2} \ll 1$.
By doing so, the ratio $\Phi\left(x\right) / \Lambda\left(x\right)$ is fixed to be a small number $\sigma_0^{2}$, which means that each observation  has the same small probability  to have its residual follow a thick-tailed distribution, and thus  an ``outlier.''
Similar to \eqref{equation_4_1_3_15}, this formulation brings the following implicit two-layer Gaussian model of $\by_n$'s {given} $\hat{\xi}_n$'s and $\bg$ during the variational inference:
\begin{equation*}
\begin{aligned}
\bbf\left(\bx_n\right) \sim \mathcal{N}\left(\bg\left(\bx_n\right),\Lambda\left(\bx_n\right)\right) \ \ \mbox{and} \ \ \by_n  \sim \mathcal{N}\left(\bbf\left(\bx_n\right),\hat{\xi}_n^{-2} \sigma_0^2 \cdot\Lambda\left(\bx_n\right)\right)  
\end{aligned}
\end{equation*}
which implies $\by_n \sim \mathcal{N}\left( \bg\left(\bx_n\right),\left(1+\hat{\xi}_n^{-2}\sigma_0^2\right)\cdot\Lambda\left(\bx_n\right)\right)$ {given} $\hat{\xi}_n$'s and $\bg$, and, moreover:
\begin{equation}
\left. \bbf\left(\bx_n\right) \mid \by_n, \bg\left(\bx_n\right) \right. \sim \mathcal{N}\left(\frac{\hat{\xi}_n^2}{\hat{\xi}_n^2+\sigma_0^2}\cdot \by_n+\frac{\sigma_0^2}{\hat{\xi}_n^2+\sigma_0^2}\cdot \bg\left(\bx_n\right),\frac{\sigma_0^2}{\hat{\xi}_n^2+\sigma_0^2}\cdot\Lambda\left(\bx_n\right)\right),
\label{equation_4_1_3_16}
\end{equation}
which means that 
the posterior mean of $\bbf\left(\bx_n\right)$, {given} the $\hat{\xi}_n$'s and $\bg$, is  a weighted average of $\by_n$ and $\bg\left(\bx_n\right)$. Thus,
the weight $\frac{\sigma_0^2}{\hat{\xi}_n^2+\sigma_0^2}$ on $\bg\left(\bx_n\right)$ can be regarded as the {\it likelihood} for $\by_n$ to be an outlier. Clearly, HeGPR-G is equivalent to having $\sigma_0 = 0$.

Here, we do not estimate $\sigma_0$ based on the likelihood of the training data as we wish to declare and exclude outliers in this model.
To tune the hyperparameter $\sigma_0$, we may adopt the Cramer-von Mises criterion for assessing multivariate normality [\cite{Koziol1982}]. To be specific, for a given $\sigma_0$ in a (finite) set of candidates $\Sigma_0$, we first estimate the model parameters, and then compute the score:
\begin{equation}
J \left(\sigma_0\right) \triangleq \frac{1}{12N}+\sum_{n=1}^{N}{\left(W_{\left(n\right)}-\frac{2n-1}{2N}\right)^{2}},
\label{equation_4_1_3_17}
\end{equation}
where $W_{\left(1\right)}\leq W_{\left(2\right)}\leq \cdots \leq W_{\left(N\right)}$ are the ordered values of the following $W_n$'s:
\begin{equation*}
W_n \triangleq F_Q\left(\left(y_n-\overline{\mu}\left(x_n\right)\right)^\top \left(\overline{\nu}\left(x_n\right)+\left(1+{\hat{\xi}}_n^{-2}\sigma_0^{2}\right)\Lambda\left(x_n\right)\right)^{-1}\left(y_n-\overline{\mu}\left(x_n\right)\right)\right),
\label{equation_4_1_3_18}
\end{equation*}
for the cumulative density function $F_Q$ of the $\chi^2$ distribution with $Q$ degrees of freedom, where $\overline{\mu}$, $\overline{\nu}$ and $\hat{\Lambda}$ are the same as those in \eqref{equation_3_4_1}. We then pick 
$\hat{\sigma_0} =\arg\min_{\sigma_0\in \Sigma_0} \min J \left(\sigma_0\right)$.

In the above formulation, after declaring that the $n$th observation is an outlier, we regard that the true $\by_n$ follows 
$\by_n \sim \mathcal{N}\left( \bg(\bx_n),(1+\hat{\xi}_n^{-2}\sigma_0^2)\cdot\Lambda(\bx_n)\right)$, which scales the precision matrix $\Lambda^{-1}\left(x_n\right)$ by $(1+\hat{\xi}_n^{-2}\sigma_0^2)^{-1}$. 
We define the average precision scaling:
\begin{equation}
\sigma_1^{-2} 
=\frac{1}{N}\sum_{n=1}^{N}\frac{{\hat{\xi}}_n^2}{{\hat{\xi}}_n^2+\sigma_0^2}.
\label{equation_4_1_3_19}
\end{equation}
Then, the posterior predictive distribution of $\by$ at a query covariate $\bx$ \textit{after nullifying outliers} can be approximated by $p (\by\mid Y;\hat{\bL},\hat{\Gamma},\hat{\mathcal{H}}) \approx \mathcal{N}(\by\mid\overline{\mu}(\bx),\overline{\nu}(\bx) + \sigma_1^{2} \cdot \hat{\Lambda}(\bx))$. The resulting method is denoted as HeGPR-O. Clearly, $\sigma_1^2 = 1$ if $\sigma_0 = 0$, thus HeGPR-G is a special case of HeGPR-O.

\subsection{Classification and state-space models}\label{sec4_2}

{\bf Classification}. Here we  focus on the binary classification problem and $Q=1$, in which one might assume that the logits of the responses follow a Gaussian process [\cite{Rasmussen2006,Hensman2015}], i.e., for a binary response $y \in \left\{0,1\right\}$ at a covariate $\bx$,
\begin{equation*}
\begin{aligned}
\Pr(y=1\mid \bx) = \frac{\exp(g(\bx))}{1+\exp\left(g\left(\bx\right)\right)},
\end{aligned}
\end{equation*}
for $g \sim \cGP\left(\bmu,\bbV \right)$. Once we have obtained (or approximated) the posterior predictive distribution $p\left(g\left(\bx\right)\middle|X,Y\right)$ for a training dataset $\left(X,Y\right)$, the posterior predictive probability of $y = 1$,
however, is not  in a closed-form and needs  to be approximated numerically.

Another setting for the Gaussian process classification considers the probits, instead of logits, of the responses to follow a Gaussian process [\cite{Liu2022}], i.e., we have:
\begin{equation}
\begin{aligned}
\Pr(y=1 \mid \bx) =\Phi\left(g(\bx)/\sqrt{\Lambda(\bx)}\right)
\end{aligned}
\label{equation_4_2_1}
\end{equation}
for a positive function $\Lambda:\mathbb{R}^P\rightarrow\mathbb{R}_{>0}$. Moreover, \eqref{equation_4_2_1} is equivalent to the following two-layer model after marginalizing $f$:
\begin{equation}
\begin{aligned}
y \sim {\rm{Bernoulli}}\left(1_{\left(-1\right)^{y}f\left(\bx\right)<0}\right) \ \ \mbox{and} \ \ f\left(\bx\right) \sim \mathcal{N}\left(g\left(\bx\right),\Lambda\left(\bx\right)\right),
\end{aligned}
\label{equation_4_2_2}
\end{equation}
thus $\Lambda$ in \eqref{equation_4_2_1} can be also understood as a variance function.

For a query covariate $\bx$, suppose that the posterior predictive distribution $p\left(f\left(\bx\right)\mid X,Y\right)$ for a training dataset $\left(X,Y\right)$ is either given or approximated by a Gaussian distribution $\mathcal{N}\left(f\left(\bx\right)\middle|\overline{\mu}\left(\bx\right),\overline{\nu}\left(\bx\right)+\hat{\Lambda}\left(\bx\right)\right)$, just as \eqref{equation_3_4_1} in Section \ref{sec3_6}. Then, the posterior predictive distribution of the corresponding label $y$ is given in a closed form as follows:
\begin{equation*}
\begin{aligned}
p\left(y=1\middle|X,Y\right)=\Phi\left(\frac{\overline{\mu}\left(\bx\right)}{\sqrt{\overline{\nu}\left(\bx\right)+\hat{\Lambda}\left(\bx\right)}}\right).
\end{aligned}
\label{equation_4_2_3}
\end{equation*}

Moreover, [\cite{Liu2022}] extend the model \eqref{equation_4_2_2} so that it also considers mislabeled responses in a given probability $1 \gg \delta > 0$:
\begin{equation}
\begin{aligned}
y \sim {\rm{Bernoulli}}\left(\left(1-\delta\right)^{1_{\left(-1\right)^{y}f\left(\bx_n\right)<0}}\cdot\delta^{1_{\left(-1\right)^{y}f\left(\bx_n\right)\geq 0}}\right),
\end{aligned}
\label{equation_4_2_4}
\end{equation}
which results in the posterior predictive distribution as follows:
\begin{equation}
\begin{aligned}
p\left(y=1\middle|X,Y\right)= \delta + \left(1-2\delta\right) \cdot
\Phi\left(\frac{\overline{\mu}\left(\bx\right)}{\sqrt{\overline{\nu}\left(\bx\right)+\hat{\Lambda}\left(\bx\right)}}\right).
\end{aligned}
\label{equation_4_2_5}
\end{equation}

Note that [\cite{Liu2022}] only considers the homoscedastic Gaussian process classification (HoGPC), i.e., $\Lambda\left(\bx\right) \equiv a$ for some positive scalar parameter $a$. Here, we can easily generalize their formulation by considering \eqref{equation_4_2_4} to be the third-level model in Section \ref{sec2_2}, i.e., to the heteroscedastic Gaussian process classification (HeGPC).



\medskip

{\bf State-space models}. A classical state-space model 
relies on a Markov structure to specify the evolution of its hidden states, and uses recursive methods such as the forward-backward algorithm \citep{Durbin1998}, Kalman filters, and particle filters \citep{Murphy2012} for computation.
In contrast, a GP state-space model \citep{Frigola2014,Eleftheriadis2017} models the hidden states by a Gaussian process (GP) so that their joint distribution is multivariate Gaussian. This brings in several advantages: 1) a GP is not necessarily Markovian and allows for long-term memory; 2) the GP model can take unevenly spaced multi-dimensional covariates; 3)  the posterior predictive distribution of the hidden state at an arbitrary covariate can be approximated well in a closed-form in many cases; 4) some  kernels (such as Matern) \citep{Rasmussen2006} correspond to solutions of certain stochastic differential equations \citep{Stein1999}, implying that one can solve such equations {indirectly} from the data.

The third-level model in equation \eqref{equation_2_2_0} of the state-space model corresponds to a pre-specified relationship between a hidden state and the associated response, thus it is usually given a priori and fixed.
Thus, the HoGP state-space model that assumes a single covariance matrix for the uncertainty of hidden states  may be too rigid, especially for cases where the covariates of the data are unevenly spaced or responses are heteroscedastic over covariates. The generalized HeGP model gives a more flexible structure on the hidden states. 

\section{Simulation Studies}\label{sec4_0}



\subsection{Regression analysis} \label{sec4_0_1}
{\bf GP regression with heteroscedastic errors.} We generate two time series with correlated noises that are heteroscedastic over time as follows: 
\begin{enumerate}
    \item  Draw a 2-dimensional mean function $\mu:\left[-5,5\right]\rightarrow\mathbb{R}^2$ as $\mu \sim \mathcal{GP}\left(0,\Sigma\otimes\mathbb{K}\right)$, where 
    $\Sigma$ is created randomly\footnote{We first generate a $2 \times 2$ matrix $A$ of which entries are i.i.d. random samples drawn from $\mathcal{N}\left(0,1^2\right)$ and then define $\Sigma$ as the correlation matrix corresponding to $A A^\top$. Throughout the paper, those described as 'random correlation matrices' are all defined by this construction.} 
    and kernel $\mathbb{K}\left(x,x'\right) = \exp{\left(-{\left|x-x'\right|}^2\right)}$; 
    \item Draw 5 independent $2 \times 2$ {correlation} matrices $\left\{V_k\right\}_{k=1}^5$ in the same way as in step 1,  and let $R\left(x\right)\triangleq \sum_{k=1}^{5} \frac{e^{-\left|x-\left(2k-6\right)\right|^2}V_k}{\sum_{l=1}^{5}e^{-\left|x-\left(2l-6\right)\right|^2}}$; 
    \item Draw $x_n \sim_{i.i.d.} U\left(-5,5\right)$ and $\by_n \sim \mathcal{N}\left(\mu\left(x_n\right),R\left(x_n\right)\right)$, $n = 1,\cdots,500$; and let $X = \left\{x_n\right\}_{n=1}^{N}$ and $Y = \left\{\by_n\right\}_{n=1}^{N}$ be our observations.
\end{enumerate}

\begin{figure}[ht]
\renewcommand{\baselinestretch}{1.0}
\centering
\includegraphics[width=0.24\textwidth] {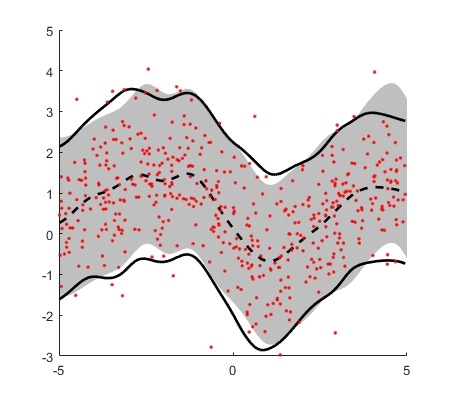}
\includegraphics[width=0.24\textwidth] {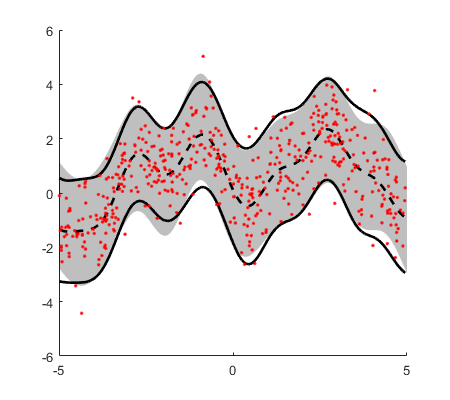}
\includegraphics[width=0.24\textwidth]  {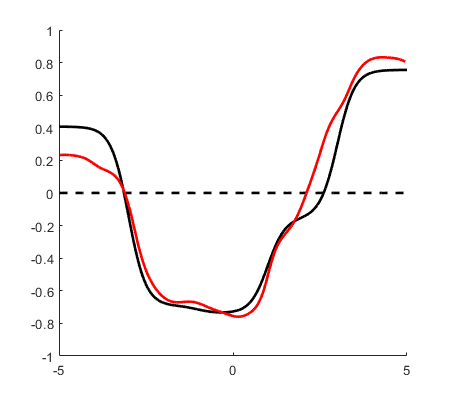}
\includegraphics[width=0.24\textwidth]  {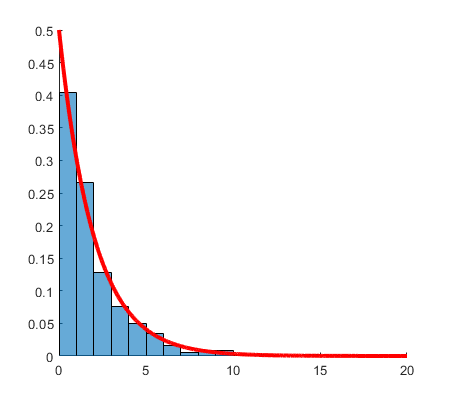}
\caption{The first and second panels are two jointly simulated time series (red dots). Shaded regions are the 95\% confidence bands based on the true generative model, while the black curves are the estimated 95\% confidence bands; the dashed curves are the  medians, respectively. The third panel shows the true (black) and estimated (red) residual correlation over covariates. The fourth panel shows the histogram of the square of the standardized residuals from the posterior predictive model overlayed with the $\chi^2{\left(2\right)}$ density (red). 
}
\label{fig_4_1_4_1}
\end{figure}

The simulated data can be viewed as bivariate responses correspond to a one-dimensional covariate, time $t$, as shown in the first and second panels of Figure \ref{fig_4_1_4_1}.
The third panel of Figure \ref{fig_4_1_4_1} compares the estimated heteroscedastic residual correlations over time with the true one, and the fourth one demonstrates that the standardarized residuals from the fitted model, ${\left(\overline{\nu}\left(\bx_n\right)+\hat{\Lambda}\left(\bx_n\right)\right)}^{-\frac{1}{2}}\left(\by_n-\overline{\mu}\left(\bx_n\right)\right)$, agrees with the postulated error model
(as the sum  of squares  of these standardized residuals are supposed to follow a Chi-squared distribution with 2 degrees of freedom if both means and covariances are properly estimated.)

We did another simulation with \textit{uncorrelated} residuals, as shown in Figure \ref{fig_H_4_1_4_1} in Appendix \ref{secH}. This example suggests that our method may overestimate
the residual correlation between multiple responses, thus constraining $\blambda_d$'s in $\bL$ to be diagonal (see Appendix \ref{secF}) or considering a nontrivial prior $\pi$ (e.g., an inverse-Wishart prior with a diagonal matrix parameter) in \eqref{equation_2_3_1} might be a practical idea to avoid it if one has a good reason to believe that errors are independent between the given multiple datasets.

\medskip

{\bf Dealing with outliers.} To 
test the effectiveness of our robust method HeGPR-O in 
Section \ref{sec4_1_4}, we simulated the observed data in the same way as  described earlier in this section except that 
(i) we only simulated 1-dimensional mean function $\mu$ from $\mathcal{GP}\left(0,\mathbb{K}\right)$ and the corresponding response $y$; (ii)
we randomly choose 5\% of the $y_i$'s  and replace them with i.i.d. draws from  
Uniform$[a,b]$, where $a$ and $b$ are   the 5\% and 95\% quantiles of $Y$, respectively. 
In this way, we generate a time series with heteroscedastic noises and also with 5\% outliers not  following the true generative model. The applying HeGPR-O model assumes the same type of kernel covariance function, but the parameters of the model will be estimated from the data.

\begin{figure}
\renewcommand{\baselinestretch}{1.0}
\centering
\includegraphics[width=0.23\textwidth,height=0.32\textwidth] {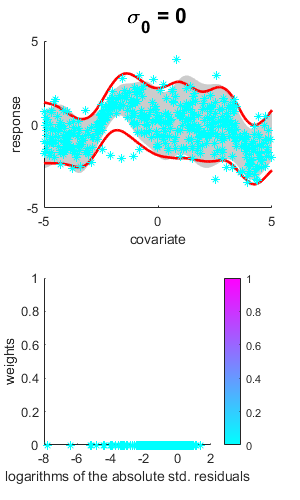}
\includegraphics[width=0.23\textwidth,height=0.32\textwidth] {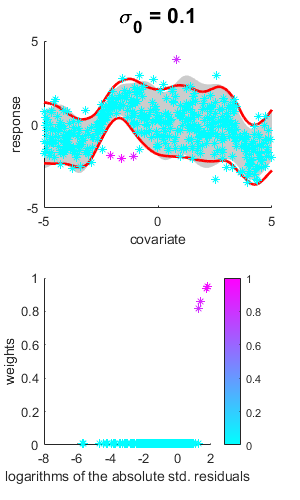}
\includegraphics[width=0.23\textwidth,height=0.32\textwidth] {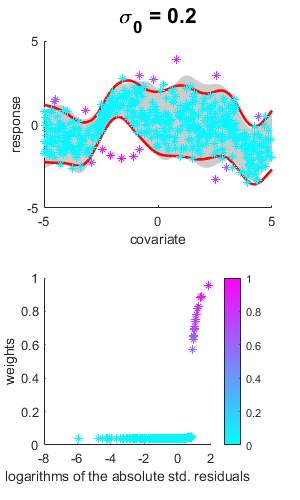}
\includegraphics[width=0.23\textwidth,height=0.32\textwidth] {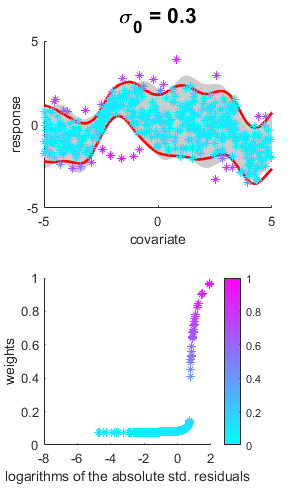}
\caption{A visualization of the estimated weights $\frac{\sigma_0^2}{\hat{\xi}_n^2+\sigma_0^2}$ in the HeGPR-O model for the heteroscedastic simulated data with outliers, for $\sigma_0 = 0,0.1,0.2,0.3$. Colors of the points are corresponding to the estimated weights indicated in the right-side color bar. (Upper) The scatter plot of the simulated responses over the associated covariates. Black regions indicate the 95\% confidence bands of the true generative model and red curves are the estimated 95\% confidence bands by the HeGPR-O model. (Lower) The scatter plot of the estimated weights over the logarithms of the absolute values of the corresponding standardized residuals.}
\label{fig_4_1_4_3}
\end{figure}

Figure \ref{fig_4_1_4_2} in Appendix \ref{secH} shows logarithms of the resulting Cramer-von Mises statistics and average KL divergences from $\mathcal{N}\left(\by\middle|\overline{\mu}\left(\bx\right),\overline{\nu}\left(\bx\right) + \sigma_1^{2} \cdot \hat{\Lambda}\left(\bx\right)\right)$ for $\sigma_1^2$ in \eqref{equation_4_1_3_19} in Section \ref{sec4_1_4} to the true generative model for $\sigma_0 = 0,0.025,0.05,\cdots,0.275,0.3$. Based on the criterion that chooses $\sigma_0$ of the smallest Cramer-von Mises statistic, we chose $\sigma_0 = 0.125$, but the minimum average KL divergence was attained at $\sigma_0 = 0.1$. Nevertheless, the patterns of two graphs that reject too small or too large $\sigma_0$'s are roughly consistent. Figure \ref{fig_4_1_4_3} visualizes the estimated weights $\frac{\sigma_0^2}{\hat{\xi}_n^2+\sigma_0^2}$ for $\sigma_0 = 0,0.1,0.2,0.3$. As expected, more responses were assigned lower weights and the estimated variances get smaller as $\sigma_0$ increases. While a too small $\sigma_0$ is ineligible for dealing with outliers, a too big $\sigma_0$ simply regards all responses of large residuals as outliers so that the model underestimates the variance of residuals.

\begin{figure}
\renewcommand{\baselinestretch}{1.0}
\centering
\includegraphics[width=0.23\textwidth] {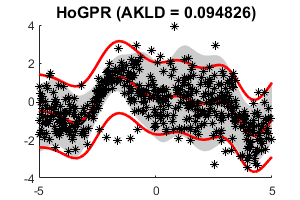}
\includegraphics[width=0.23\textwidth] {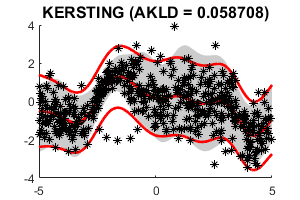}
\includegraphics[width=0.23\textwidth] {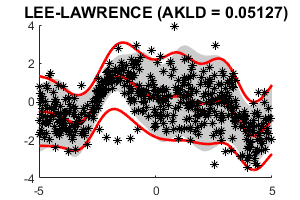}
\includegraphics[width=0.23\textwidth] {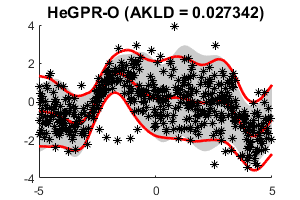}
\caption{Regression results for various Gaussian process regression models on the simulated data with heteroscedastic errors and outliers. In each panel, the black region indicates the 95\% confidence bands of the true generative model and red curves are the estimated 95\% confidence bands. Black stars are the responses. The average Kullback-Leibler divergences (AKLD) from the regression models to the true generative model are in the brackets of the titles.}
\label{fig_4_1_4_4}
\end{figure}

We next compared our HeGPR-O model of $\sigma_0 = 0.125$ with the homoscedastic Gaussian process regression model (HoGPR, which is a special case of HeGP by constraining $\blambda_d \equiv \blambda_0$ for a single variance parameter $\blambda_0$), the previous heteroscedastic Gaussian process regression model \textit{without} the outlier modeling step \citep{Lee2019} (LEE-LAWRENCE) and the method of 
\citep{Kersting2007} (KERSTING), as shown in Figure \ref{fig_4_1_4_4}. Though LEE-LAWRENCE and KERSTING show better performance than HoGPR as their average KL divergences are 0.0513 and 0.0587 bigger than 0.0948 of HoGPR, our HeGPR-O with that of 0.0273 worked best.

We did another simulation of HeGPR-O on the simulated data \textit{without} outliers, as shown in Figures \ref{fig_H_4_1_4_2} and \ref{fig_H_4_1_4_3} in Appendix \ref{secH}. This example suggests that our criterion might overestimate $\sigma_0$ (if all residuals are supposed to be Gaussian, then the proper $\sigma_0 = 0$) but the regression model is insensitive to the choice of $\sigma_0$.

\subsection{A classification simulation} \label{sec4_0_2}
\begin{figure}[ht]
\renewcommand{\baselinestretch}{0.9}
\centering
\includegraphics[width=0.24\textwidth]{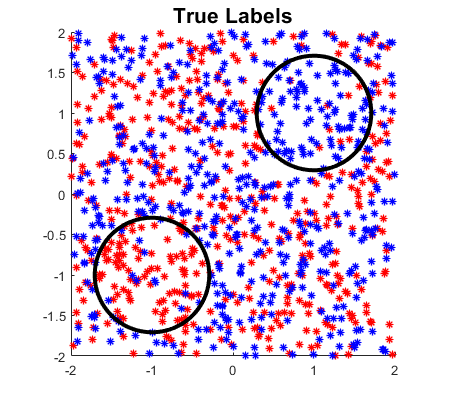}
\includegraphics[width=0.24\textwidth]{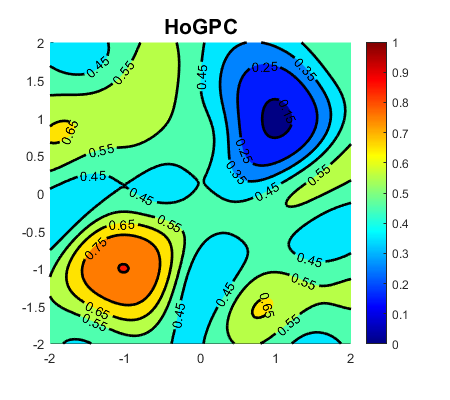}
\includegraphics[width=0.24\textwidth]{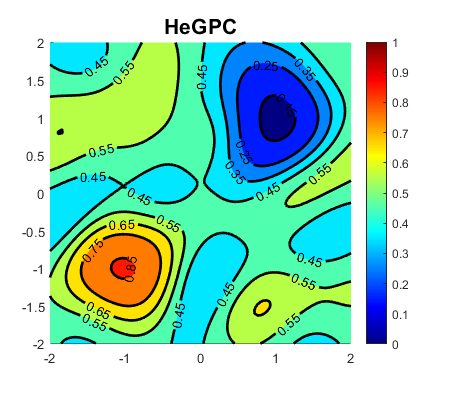}
\includegraphics[width=0.24\textwidth]{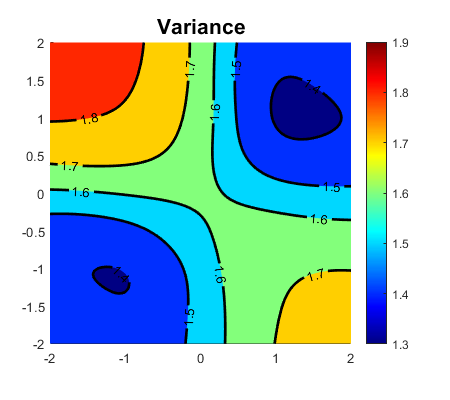}
\caption{Classification results for a simulated dataset by using (second) the HoGP model, and (third) the HeGP model. The simulated red-blue categories with true boundaries are shown in the first panel. In each circle, firstly a red or blue category is sampled with a 100\% chance and 50-50 chance outside of the circles, and then we randomly permute them in a probability of 10\%. The estimated heteroscedastic variance $\Lambda$ is visualized in the fourth panel. Here, we chose the true $\delta=0.1$ and kernel was the squared-exponential kernel. The average KL divergences of the true label probability from those of HoGPC and HeGPC models are 0.025257 and 0.024461, respectively.}
\label{fig_4_2_1}
\end{figure}

We  generated 1000 points uniformly in the square $\left(-2,2\right)\times\left(-2,2\right)$, as shown in Figure \ref{fig_4_2_1} (upper left). The points are labeled as red and blue independently with  probability 0.5 for those outside the two circles (centered at $\left(-1,-1\right)$ and $\left(1,1\right)$), and with probabilities (0.95, 0.05) and (0.05, 0.95) for those in each circle, respectively.  

We fit the GP-based probit model \eqref{equation_4_2_2} for both the original HoGPC setting of \cite{Liu2022} and our HeGPC setting with non-constant precision process $\Lambda^{-1}(x)$.
Figure \ref{fig_4_2_1} shows a comparison between the results based on  HoGPC and HeGPC models, respectively, for the simulated dataset, where $\delta=0.1$ in \eqref{equation_4_2_4} is given {\it a priori}. Both models detect regions that are biased to one category, and the two models differ very minimally.  HeGPC model seems to infer more a bit more variability in the 50-50 regions
and less variability in biased regions.
This difference is more vivid in Figure \ref{fig_H_4_2_2} in Appendix \ref{secH}, where only a small portion of the training dataset are in the 50-50 region.

 \subsection{A state-space model}\label{sec4_0_3}

We simulated a HeGP state-space model as follows: 

\begin{enumerate}
    \item draw a mean function $\bmu:\mathbb{R} \rightarrow \mathbb{R}^{3}$ from a Gaussian process $\mathcal{GP}\left(0,\Sigma\otimes\mathbb{K}\right)$ for a squared-exponential kernel $\mathbb{K}\left(x,x'\right) \triangleq \exp \left(-2\left(x-x'\right)^2\right)$ and $\Sigma \triangleq \frac{1}{9}\Sigma_0$ for a random $3 \times 3$ \textit{correlation} matrix $\Sigma_0$;
    \item draw five $3 \times 3$ random matrices $\left\{U_k\right\}_{k=1}^5$ where each entry $u_k(i,j)\sim \text{Unif}(0,0.3)$, and let $V_k \triangleq U_kU_k^\top$. Define $\Omega\left(x\right)\triangleq{\sum_{k=1}^{5}{e^{-\left|x-\left(1.5k-4.5\right)\right|^2}V_k}}/{\sum_{k=1}^{5}e^{-\left|x-\left(1.5k-4.5\right)\right|^2}}$, and let the true generative model be $\bbf\left(x\right) \sim \mathcal{N}\left(\bmu\left(x\right),\Omega\left(x\right)\right)$ for $x \in \mathbb{R}$;
    \item draw hidden signals $F = \left\{\bbf\left(x_n\right)\right\}_{n=1}^{400}$ from the generative model for $X = \left\{x_n\right\}_{n=1}^{400}$ that are regularly spaced over $\left[-3,3\right]$;
    \item draw 3-dimensional observed signals $Y=\left\{\by_n\right\}_{n=1}^{400}$ given $F$ as follows:
    \begin{equation}
\begin{aligned}
\by_{n1} & \triangleq \left[\bbf\left(x_n\right)\right]_1+0.05\cdot\epsilon_{n1}, \quad \epsilon_{n1} \sim \mathcal{T}\left(6\right) \\ \by_{n2} & \triangleq \exp\left(\left[\bbf\left(x_n\right)\right]_2\right)+\left[\bbf\left(x_n\right)\right]_2+0.10\cdot\epsilon_{n2}, \quad \epsilon_{n2} \sim \mathcal{T}\left(6\right) \\ \by_{n3} & \triangleq \exp\left(\frac{1}{2}\left[\bbf\left(x_n\right)\right]_3\right)+\left(\left[\bbf\left(x_n\right)\right]_3\right)^3+0.15\cdot\epsilon_{n3}, \quad \epsilon_{n2} \sim \mathcal{T}\left(6\right).
\end{aligned}
\label{equation_4_3_1}
\end{equation}
\end{enumerate}

\begin{figure}[ht]
\renewcommand{\baselinestretch}{1}
\includegraphics[width=0.24\textwidth]{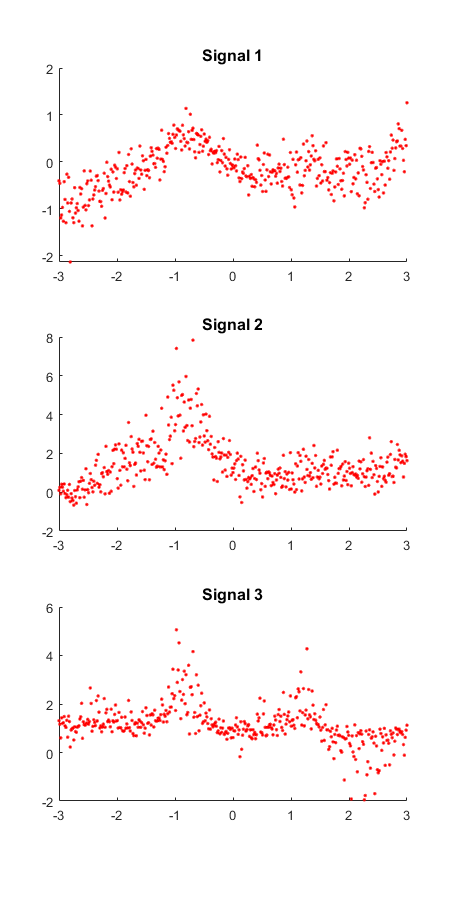}
\includegraphics[width=0.24\textwidth]{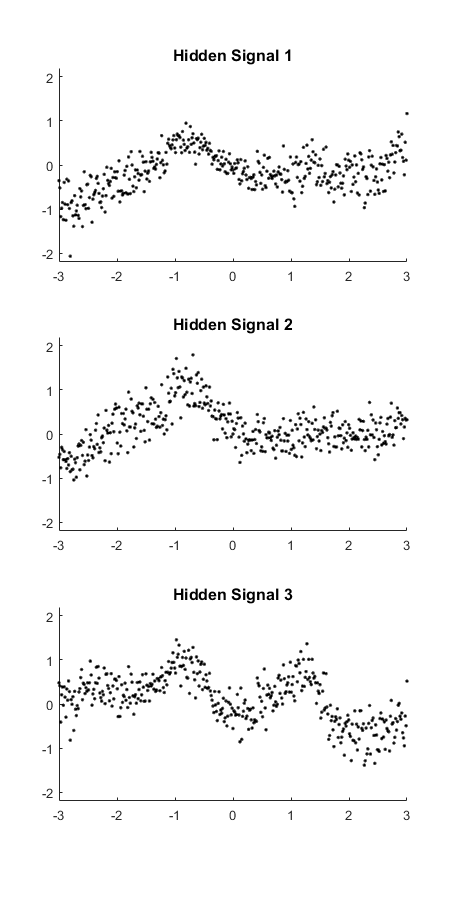}
\includegraphics[width=0.24\textwidth]{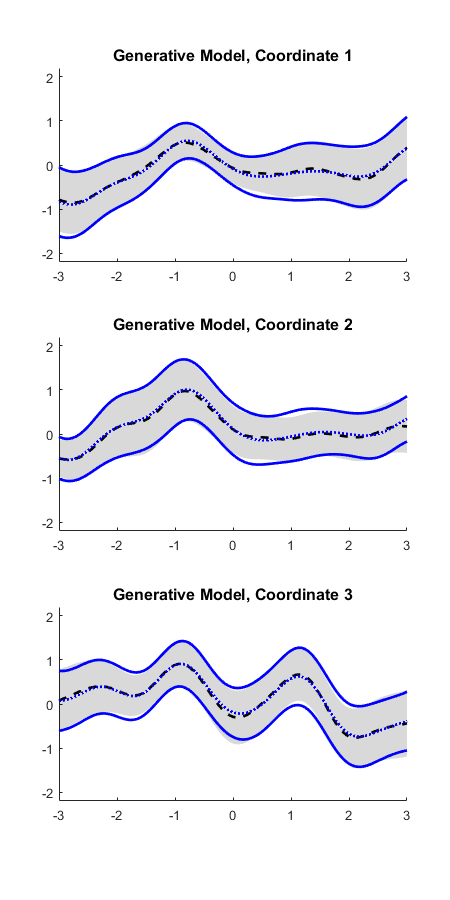}
\includegraphics[width=0.24\textwidth]{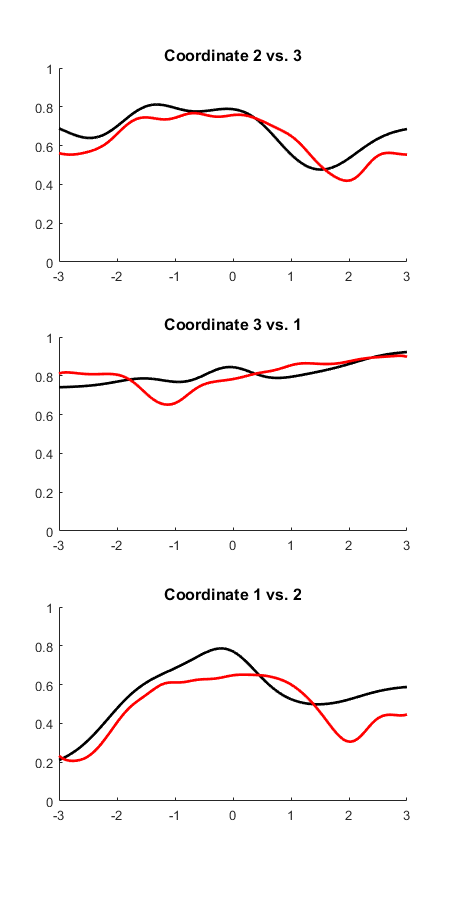}
\caption{Column 1: simulated signals $Y$;
Column 2: the underlying hidden states $\bff_X$; 
Column 3: the inferred generative model (dotted curves)  with the 
$\pm 1.96\sigma_q(x)$ bands (blue curves), $q=1,2,3$,  covering 95\% of the generated $f_q(x)$, overlaid by true generative model (dashed curves) and its 95\% probability band (dark shaded regions);
Column 4: estimated residual correlations (red) overlaid by the true ones (black).}
\label{fig_4_3_1}
\end{figure}

Figure \ref{fig_4_3_1} (columns 1 and 2) shows the simulated signals $Y$ and true hidden signals $F$. We run the HeGP state-space model inference on $\left(X,Y\right)$ \textit{given} the relationship \eqref{equation_4_3_1}, but $\bmu$, $\Omega$, $\Sigma$, and kernel hyperparameters of $\mathbb{K}$ are to be estimated.  The inferred generative model $\bbf$ together with the comparison between the true and estimated residual correlations between coordinates are shown in columns 3 and 4 of Figure \ref{fig_4_3_1}, respectively.  Figure \ref{fig_4_3_3} of Appendix \ref{secH} shows  the chi-squared statistics of the standardized residuals of the true hidden signals with respect to both the true generative model and  the estimated one,  overlaid with the  density function of the $\chi^2$(3) distribution. The estimated hidden signal $F$ is shown in Figure \ref{fig_H_4_3_5} of Appendix \ref{secH}. 



\section{Applications to Climatology}\label{sec5}


\subsection{Construction of the UK37 and TEX86 calibration curves}\label{sec5_1}

In paleoceanography and paleoclimatology, the sea surface temperature (SST) reconstruction usually depends on a few relevant proxies, such as ${\rm{U}}_{37}^{\rm{K}\prime}$ [\cite{TIERNEY2018}] and TEX\textsubscript{86} [\cite{Kim2010}], under the assumption that the relevance between SST and a proxy is consistent over time. To be more specific, let $x = x\left(t\right)$ be an unknown SST that is associated with a proxy observation $y$ at time $t$. Then, the inference can be done by a prior on $x$ and the likelihood of $y$ given $x$. While the prior is often based on the spatial and chronological information of the SSTs associated with $y$, the likelihood of $y$ given $x$ depends on a (given) calibration model, which is derived from a set of present-day values of SST and proxy pairs. Thus, constructing a reliable calibration model $p\left(y\middle| x\left(t\right)\right)$ is important in this field of research.

For the construction of UK37 proxy calibration model, \cite{TIERNEY2018} adopt a Bayesian B-spline regression model and \cite{Lee2019} apply a HeGP regression to the data. Moreover, \cite{Tierney2014} employ a HoGP regression for constructing a TEX86 calibration model. Here we applied our HeGPR-O model in Section \ref{sec4_1_4} to the construction of calibration models of TEX\textsubscript{86} and ${\rm{U}}_{37}^{\rm{K}\prime}$. The same TEX\textsubscript{86} and ${\rm{U}}_{37}^{\rm{K}\prime}$ proxy values with those of \cite{Tierney2014} and \cite{TIERNEY2018}, respectively, were used as the training datasets. Since both proxies are constrained to $\left[0,1\right]$, we first applied the logit transformation $y \rightarrow \log{\left(y/\left(1-y\right)\right)}$ to the proxy values (responses) and then standardized by its mean and standard deviation for each transformed proxy, before running the algorithm: values in the ranges $\left[0,10^{-10}\right]$ and $\left[1-10^{-10},1\right]$ are identified as $10^{-10}$ and $1-10^{-10}$, to avoid $-\infty$ and $\infty$ in the logit transformations, respectively. SSTs are also standardized by $x\rightarrow\left(x-16\right)/8$. We set the induced SSTs (covariates) $\underline{X} = X$, the adjacent percentage parameter ${A}=5$ and candidates $\mathcal{R}=\left\{1,1.5,2,2.5,\cdots,20\right\}$ for the percentage parameter discussed in Appendix \ref{secB1}. The density kernel $\mathcal{K}$ was a Gaussian kernel. Four steps in Algorithm \ref{algorithm1} were iterated for 300 times, and for each iteration, 100 iterations were applied to the gradient ascent in E-step. 

For the Gaussian process modeling on $\bg$ in \eqref{equation_2_0}, we chose the Matérn covariance kernel $\mathbb{K}$ with unknown kernel hyperparameters $\sigma$ and $\gamma$ as follows:
\begin{equation*}
\mathbb{K}\left(x,x'\right) \triangleq\sigma^2 \left(1+\sqrt3\gamma^2\left|x-x'\right|\right)\exp\left(-\sqrt3\gamma^2\left|x-x'\right|\right),
\label{equation_5_1_1}
\end{equation*}
and the mean function $\mu\left(x\right) \triangleq ax+b$ for scalar parameters $a$ and $b$.

To determine $\sigma_0$, we compute logarithms of the resulting Cramer-von Mises statistics for $\sigma_0 \in \left\{0,0.025,0.05,\cdots,0.3\right\}$, as shown in Figure \ref{fig_5_1_1_0} in Appendix \ref{secH}. Based on the criterion discussed in Section \ref{sec4_1_4}, we chose $\sigma_0=0.075$ for TEX\textsubscript{86} and $\sigma_0=0.125$ for ${\rm{U}}_{37}^{\rm{K}\prime}$.

\begin{figure}[ht]
\renewcommand{\baselinestretch}{1}
\includegraphics[width=0.24\textwidth]{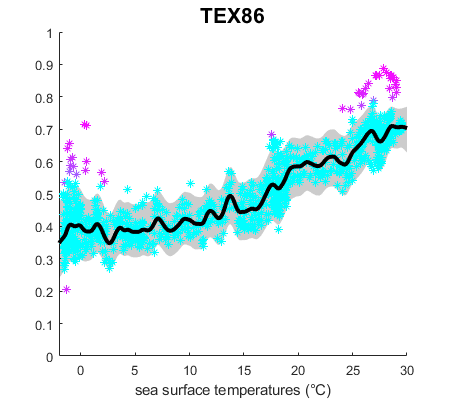}
\includegraphics[width=0.24\textwidth]{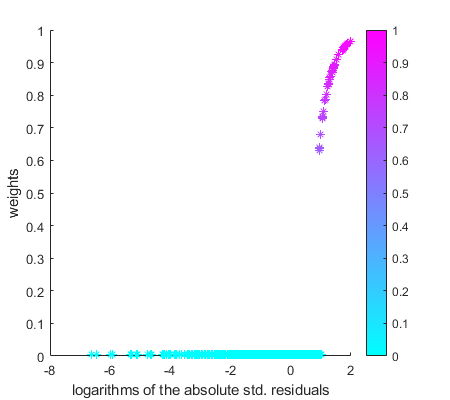}
\includegraphics[width=0.24\textwidth]{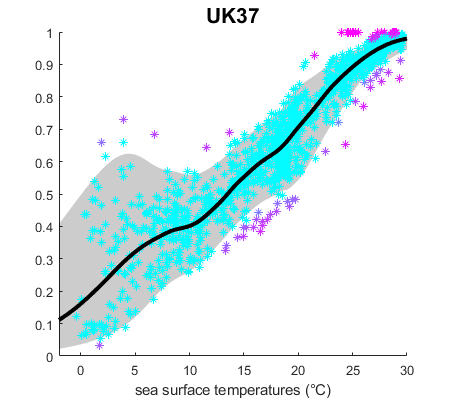}
\includegraphics[width=0.24\textwidth]{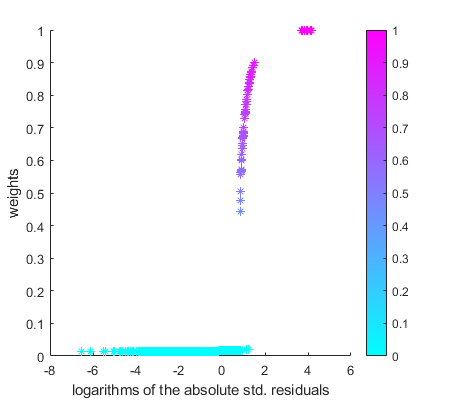}
\caption{In all panels, the color scheme of the points corresponds to the values of the estimated  weights $\frac{\sigma_0^2}{\hat{\xi}_n^2+\sigma_0^2}$'s (as shown the right-side color bar). The first and third panels correspond to the HeGPR-O calibration model for TEX\textsubscript{86} and ${\rm{U}}_{37}^{\rm{K}\prime}$, respectively. The shaded region is the 95\% confidence band of the calibration model, and the black curve is medians of the posterior predictive distribution of the underlying hidden function $g\left(x\right)$. The second and fourth panels correspond to the scatter plot of the estimated $\frac{\sigma_0^2}{\hat{\xi}_n^2+\sigma_0^2}$'s over the logarithms of the absolute values of the corresponding standardized residuals.}
\label{fig_5_1_1_1}
\end{figure}

Figure \ref{fig_5_1_1_1}  show the corresponding regression (or calibration) models for TEX\textsubscript{86} and ${\rm{U}}_{37}^{\rm{K}\prime}$, after restoring the logits by $z \rightarrow \left({1+\exp\left(-z\right)}\right)^{-1}$, and the estimated weights $\frac{\sigma_0^2}{\hat{\xi}_n^2+\sigma_0^2}$ that quantify the magnitude of being outliers, respectively. Clearly, the variances of residuals from regression functions are heteroscedastic over SSTs for both proxies. In both regression models, apparent outliers (e.g., ${\rm{U}}_{37}^{\rm{K}\prime}$ proxy values at around \SI{25}{\degreeCelsius} above the 95\% confidence band) are successfully ignored but the inferred 95\% confidence bands cover 93.24\% of TEX\textsubscript{86} and 90.54\% of ${\rm{U}}_{37}^{\rm{K}\prime}$ proxy observations, which implies either outliers are just considerably many or there exist some clusters of data (e.g., a cluster above the regression model of TEX\textsubscript{86} at \SI{25}{\degreeCelsius} - \SI{29}{\degreeCelsius}) that cannot be explained by one calibration model over one covariate, i.e., SSTs, for each dataset, which is supported by the histograms in Figure \ref{fig_H_5_1_1}, Appendix \ref{secH}. 

\subsection{Weekly average temperatures of six US airports}\label{sec5_3}

As an example of the \textit{multivariate} HeGPR model to the real dataset, we run the HeGPR-O model in Section \ref{sec4_1_4} on the weekly average temperatures of six US airports (BOS, DCA, JFK, LAX, SEA and SFO, which stand for the Boston Logan International Airport, Ronald Reagan Washington National Airport, John F. Kennedy International Airport, Los Angeles International Airport, Seattle-Tacoma International Airport and San Francisco International Airport, respectively). To be specific, each time series ranges from January 1, 2018 to January 1, 2023 and NOAA daily average temperatures (TAVG) of each airport are averaged into a weekly average temperature from every Monday to Sunday. In this example, we are more interested in estimating the time-varying correlation of the \textit{residuals} across the airports rather than that of the regression models, which is mostly depending on the seasonal change in the northern hemisphere so somewhat trivial.

In this example, we set $\mu\left(x\right) \triangleq \Vec{b}$ for a $6 \times 1$ vector parameter $\Vec{b}$ and $\bbV \triangleq \Sigma \otimes \bbK$ for a $6 \times 6$ covariance matrix parameter $\Sigma$ and a squared-exponential kernel $\bbK\left(t,s\right) \triangleq \exp{\left(-\gamma^2\left|t-s\right|^2\right)}$, where $\gamma$ is a scalar kernel hyperparameter, for the Gaussian process prior. Here, the induced covariates $\underline{X} = X$, the adjacent percentage parameter $A=5$, and candidates $\mathcal{R} = \left\{1,1.5,2,\cdots,19.5,20\right\}$ for the percentage parameter discussed in Appendix \ref{secB1}. The density kernel $\mathcal{K}$ was a Gaussian kernel. Four steps in Algorithm \ref{algorithm1} were iterated for 300 times, and for each iteration, 100 iterations were applied to the gradient ascent in E-step. We choose $\sigma_0 = 0.1$ based on the criterion discussed in Section \ref{sec4_1_4}, as shown in Figure \ref{fig_5_3_1} in Appendix \ref{secH} for the logarithms of the Cramer-von Mises statistics for $\sigma_0 \in \left\{0,0.025,0.05,\cdots,0.275,0.3\right\}$.

\begin{figure}[ht]
\renewcommand{\baselinestretch}{1}
\centering
\includegraphics[width=1.00\textwidth]{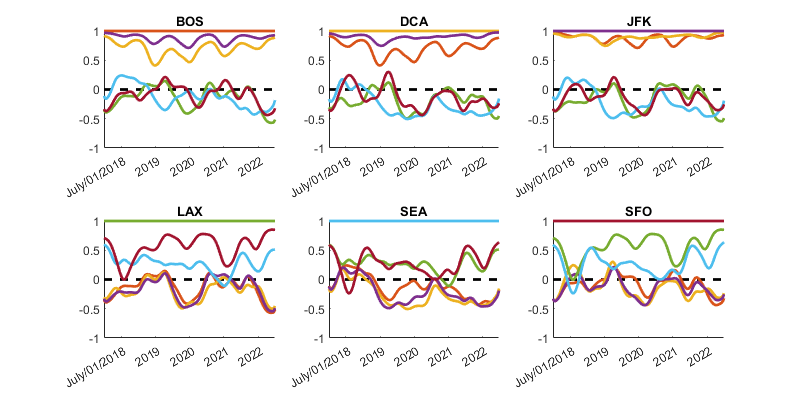}
\caption{The inferred time-varying correlation of residuals. Each year mark indicates July 1st of that year.}
\label{fig_5_3_2}
\end{figure}

Figure \ref{fig_5_3_1} in Appendix \ref{secH} shows the resulting marginal regression models for each airport. The estimated 95\% confidence bands cover 95.4\%, 93.1\%, 94.3\%, 93.9\%, 95.4\% and 92.3\% of the weekly average temperatures of BOS, DCA, JFK, LAX, SEA and SFO, respectively, demonstrating clear seasonal patterns. The estimated time-varying pairwise correlations across residuals are shown in Figure \ref{fig_5_3_2}. 
As expected from the geographical characteristics of the six airports, the time-varying correlations among BOS, DCA and JFK, and that between LAX and SFO
are more prominent than the others, and all show the seasonality.

The approximated posterior predictive distribution \eqref{equation_4_1_3_16} standardizes each 6-dimensional residual by ${\left(\overline{\nu}\left(x_n\right)+\sigma_1^2\cdot\hat{\Lambda}\left(x_n\right)\right)}^{-\frac{1}{2}}\left(y_n-\overline{\mu}\left(x_n\right)\right)$. The first panel of Figure \ref{fig_H_5_3_3} in Appendix \ref{secH} is the histogram of squares of the norms of those standardized residuals, which fit to the probability density function of the Chi-squared distribution with 6 degrees of freedom. The second panel represents the scatter plot of logarithms of the norms of standardized residuals and the estimated weights $\frac{\sigma_0^2}{\hat{\xi}_n^2+\sigma_0^2}$'s in Section \ref{sec4_1_4}, which implies that, except for five tuples, all of weekly average temperatures were almost equally treated in inference.

\section{Discussion and Conclusion}\label{sec6}

The proposed generalized HeGP model offers several distinct advantages over previously published methods. First of all, it boosts broader applicability, seamlessly accommodating a wide range of Gaussian process models without constraints on the nature of response models. Further setting it apart, the model is grounded in a variational EM framework featuring a closed-form M-step update for heteroscedastic covariance matrices, which greatly speed up the computation. 
The model's robustness extends across dimensions,  accommodating various covariates and responses. It derives variance estimates directly from the estimation process, obviating the need for logarithmic transformations employed in popular time series models such as the multivariate GARCH model,
and thereby sidestepping potential structural biases. 
Moreover, our model's computation of heteroscedastic variance at any query covariate is also direct and unequivocal.


However, we still have challenges to be tackled in future works. First, the time complexity of the exact model is $\mathcal{O} \left( N^3 Q^3 \right)$ and that of the variational free energy (VFE) approximation in Appendix \ref{secG} is still $\mathcal{O} \left( N M^2 Q^3 \right)$, where $N$ is the size of data, $M$ is that of the induced covariates, and $Q$ is the dimensionality of responses. Thus, the model quickly becomes intractable as $Q$ grows. Second, constructing an optimal ``imputed covariates'' $\underline{X}$ in Section \ref{sec2_2} and choosing an optimal size $D$ of $\underline{X}$ are still an unresolved issue -- letting $\underline{X} \equiv X$ could perform well but the corresponding time complexity is quadratic to the size of dataset, thus limited; Intuitive constructions such as evenly-spaced covariates over $X$ are usually working only for low dimensionality of covariates $P$; K-means clustering on $X$ with dimensionality reduction techniques could be considered for large $P$ in practice, but currently we do not have a definite answer to this issue. Third, though our derivation is justified by an EM algorithm, it also depends on the posterior approximation in general, which may involve  large gaps in special cases. Conditions that make that approximation accurate enough to guarantee the convergence should be specified theoretically. 


Nonetheless, our model is supported by good performance on both the simulation and real examples, as described in Sections \ref{sec4_0} and \ref{sec5}. In regression analysis, a hybrid modeling of the Student’s t-distribution and  
HeGP prior not only captures heteroscedastic error covariance over covariates but also makes the regression model robust to outliers. In classification, the HeGP prior makes the classification more certain at the regions of covariates of certain categories and less certain at that of uncertain categories. In state-space models, the posterior predictive distribution of the hidden variables reflects the heteroscedasticity of the generative models. 
It will be a good challenge to extend the derivation of our generalized HeGP model designed  to others, such as Gaussian process latent variable models [\cite{Lawrence2005,Damianou2016}] and deep Gaussian process models [\cite{Damianou2013}].

\section*{Acknowledgements}
This work was supported in part by NSF under Grant DMS-2015411; and NIH under Grant R01 HG011485-01.
The authors report there are no competing interests to declare.

\clearpage
\newpage

\appendixpage
\appendix
\section{Details of the VEM algorithm}
\label{secA}
\subsection{The E-step}\label{secA_1}
As mentioned in Section \ref{sec3_1}, in the E-step we approximate the exact marginal posterior $p (\bbf_X \mid X, {Y};\bL^{(t)},\Theta^{(t)},\Upsilon^{(t)} )$ with $q(\bbf_X \mid\ \Gamma )$ in equations \eqref{equation_3_1_6} for approximating the following Q-function in \eqref{equation_3_1_3} by \eqref{equation_3_1_9}:
\begin{equation}
\begin{aligned}
\mathcal{Q} \left(\bL,\Theta,\Upsilon\mid\bL^{\left(t\right)},{\Theta}^{\left(t\right)},{\Upsilon}^{\left(t\right)}\right) & \triangleq \mathbb{E}_{p\left(\bff_X,\bg_X\middle| X, {Y};\bL^{\left(t\right)},{\Theta}^{\left(t\right)},{\Upsilon}^{\left(t\right)}\right)}\left[\log{p\left({Y},\bff_X,\bg_X,\bL \middle| X, {\Theta},{\Upsilon}\right)}\right] \\ & \approx \mathbb{E}_{q\left(\bff_X,\bg_X\middle| \bL^{\left(t\right)},{\Upsilon}^{\left(t\right)},\Gamma\right)}\left[\log{p\left({Y},\bff_X,\bg_X,\bL \middle| X, {\Theta},{\Upsilon}\right)}\right],
\end{aligned}
\label{equation_A_1_1}
\end{equation}
by estimating the variational parameter $\Gamma = (\eta,\Psi)$ so that the KL divergence 
in \eqref{equation_3_1_7} is minimized, which is equivalent to maximizing an evidence lower bound (ELBO) $\mathcal{L}\left(\Gamma\right)$ of the log-marginal likelihood $\log{p\left(Y\middle|X;\bL^{\left(t\right)},\Theta^{\left(t\right)},\Upsilon^{\left(t\right)}\right)}$:
\begin{equation}
\begin{aligned}
\mathcal{L}\left(\Gamma\right) & \triangleq {\mathbb{E}_{q\left(\bbf_X\middle|\Gamma\right)}\left[\log{\frac{p\left(Y\middle| \bbf_X;\Theta^{\left(t\right)}\right)p\left(\bbf_X\middle|\bL^{\left(t\right)},\Upsilon^{\left(t\right)}\right)}{q\left(\bbf_X\middle|\Gamma\right)}}\right]} \\ & = \mathbb{E}_{q\left(\bbf_X\middle|\Gamma\right)}\left[\log{p\left(\bbf_X\middle|\bL^{\left(t\right)},\Upsilon^{\left(t\right)}\right)} - \log{q\left(\bbf_X\middle|\Gamma\right)}\right] + \mathbb{E}_{q\left(\bbf_X\middle|\Gamma\right)}\left[\log{p\left(Y\middle| \bbf_X;\Theta^{\left(t\right)}\right)}\right]
\end{aligned}
\label{equation_A_1_2}
\end{equation}

In \eqref{equation_A_1_2}, the first expectation is expressed in the following closed form because both $p\left(\bbf_X\middle|\bL^{\left(t\right)},\Upsilon^{\left(t\right)}\right)$ and $q\left(\bbf_X\middle|\Gamma\right)$ are Gaussian: 
\begin{equation*}
\begin{aligned}
& \mathbb{E}_{q\left(\bbf_X\middle|\Gamma\right)}\left[\log{p\left(\bbf_X\middle|\bL^{\left(t\right)},\Upsilon^{\left(t\right)}\right)} - \log{q\left(\bbf_X\middle|\Gamma\right)}\right] \\ = & -\frac{1}{2}\cdot{\vvec\left(\eta_X-\mu_X^{\left(t\right)}\right)}^\top\left(\mathbb{V}_{XX}^{\left(t\right)}+\Lambda_{XX}^{\left(t\right)}\right)^{-1}\vvec\left(\eta_X-\mu_X^{\left(t\right)}\right) \\ & - \frac{1}{2} \cdot trace\left(\left(\mathbb{V}_{XX}^{\left(t\right)}+\Lambda_{XX}^{\left(t\right)}\right)^{-1}\Psi_{XX}\right) + \frac{1}{2}\log{\left|\Psi_{XX}\right|} + Const.,
\end{aligned}
\label{equation_A_1_3}
\end{equation*}
and we can also represent its partial derivative with respect to $\Gamma$ in a closed form for the gradient ascent.

However, the second expectation might not be expressed in a closed form unless the third-level model $p\left(Y\mid \bbf_X,\Theta^{\left(t\right)}\right)$ is Gaussian. Thus, for the estimation of $\Gamma$, we apply the reparameterization trick [\cite{Kingma2013}] to the second expectation for approximating its partial derivative with respect to $\Gamma$ for feeding it into the gradient ascent.

Once the optimization is over, $\mathcal{Q}(\bL,\Theta,\Upsilon\mid\bL^{\left(t\right)},{\Theta}^{\left(t\right)},{\Upsilon}^{\left(t\right)})$ in equation \eqref{equation_A_1_1} is approximated with $\tilde{\mathcal{Q}}(\bL,\Theta,\Upsilon\mid\bL^{\left(t\right)},{\Theta}^{\left(t\right)},{\Upsilon}^{\left(t\right)})$ as follows:
\begin{equation}
\begin{aligned}
\tilde{\mathcal{Q}} \left(\bL,\Theta,\Upsilon\mid\bL^{\left(t\right)},{\Theta}^{\left(t\right)},{\Upsilon}^{\left(t\right)}\right) \triangleq \mathbb{E}_{q\left(\bff_X,\bg_X\middle| \bL^{\left(t\right)},{\Upsilon}^{\left(t\right)},\hat{\Gamma}^{\left(t\right)}\right)}\left[\log{p\left({Y},\bbf_X,\bg_X,\bL \middle|X; {\Theta},\Upsilon\right)}\right],
\end{aligned}
\label{equation_A_1_10}
\end{equation}
for the estimated variational parameter $\hat{\Gamma}^{(t)}$ as the result of optimization.

\subsection{The M-step}\label{secA_2}
The goal of the M-step is to update $\Theta^{(t+1)}$, $\Upsilon^{(t+1)}$ and $\bL^{(t+1)}$ so that they maximize the Q-function in \eqref{equation_3_1_3}. Here, we instead maximize the approximated Q-function in \eqref{equation_A_1_10}, which can be rewritten as follows:
\begin{equation*}
\begin{aligned}
& \tilde{\mathcal{Q}} \left(\bL,\Theta,\Upsilon\mid\bL^{\left(t\right)},{\Theta}^{\left(t\right)},{\Upsilon}^{\left(t\right)}\right) \\ = & \ \mathbb{E}_{q\left(\bbf_X\middle|\hat{\Gamma}^{\left(t\right)}\right)}\left[\log{p\left(Y\middle| \bbf_X;\Theta\right)}\right] \\ + & \ \mathbb{E}_{q\left(\bff_X,\bg_X\middle| \bL^{\left(t\right)},{\Upsilon}^{\left(t\right)},\hat{\Gamma}^{\left(t\right)}\right)}\left[\log{p\left(\bg_X\middle|\Upsilon\right)}\right] \\ + & \ \mathbb{E}_{q\left(\bbf_X,\bg_X\middle|\bL^{\left(t\right)},\Upsilon^{\left(t\right)},\hat{\Gamma}^{\left(t\right)}\right)}\left[\log{p\left(\bbf_X\middle| \bg_X;\bL\right)} + \log{p\left(\bL\middle|X\right)} \right],
\end{aligned}
\label{equation_A_2_1}
\end{equation*}
i.e., the update is done {\it{separately}}:
\begin{equation}
\begin{aligned}
\Theta^{\left(t+1\right)} & \triangleq \argmax_{\Theta}{\mathbb{E}_{q\left(\bbf_X\middle|\hat{\Gamma}^{\left(t\right)}\right)}\left[\log{p\left(Y\middle| \bbf_X;\Theta\right)}\right]} \\ \Upsilon^{\left(t+1\right)} & \triangleq \argmax_{\Upsilon}{\mathbb{E}_{q\left(\bff_X,\bg_X\middle| \bL^{\left(t\right)},{\Upsilon}^{\left(t\right)},\hat{\Gamma}^{\left(t\right)}\right)}\left[\log{p\left(\bg_X\middle|\Upsilon\right)}\right]} \\ \bL^{\left(t+1\right)} & \triangleq \argmax_{\bL}{\mathbb{E}_{q\left(\bbf_X,\bg_X\middle|\bL^{\left(t\right)},\Upsilon^{\left(t\right)},\hat{\Gamma}^{\left(t\right)}\right)}\left[\log{p\left(\bbf_X\middle| \bg_X;\bL\right)} + \log{p\left(\bL\middle|X\right)} \right]}.
\end{aligned}
\label{equation_A_2_2}
\end{equation}

In general, the update of ${\Theta}^{\left(t+1\right)}$ cannot be expressed in a closed form, so we resort to numerical optimization methods such as gradient ascent. To be specific, we rely on a numerical integration of $\mathbb{E}_{q\left(\bbf_X\middle|\hat{\Gamma}^{\left(t\right)}\right)}\left[\log{p\left(Y\middle| \bbf_X;\Theta\right)}\right]$:
\begin{equation}
\mathbb{E}_{q\left(\bbf_X\middle|\hat{\Gamma}^{\left(t\right)}\right)}\left[\log{p\left(Y\middle| \bbf_X;\Theta\right)}\right]\approx\frac{1}{M}\sum_{m=1}^{M}\log{p\left(Y\middle|{\widetilde{\bbf}}_X^{\left(m\right)};\Theta\right)},
\label{equation_A_2_4}
\end{equation}
for independent and identically distributed samples $\left\{{\widetilde{\bbf}}_X^{\left(m\right)}\right\}_{m=1}^M$ drawn from $q\left(\bbf_X\middle|\hat{\Gamma}^{\left(t\right)}\right)$. Note that we do not need the reparameterization trick here because $\hat{\Gamma}^{\left(t\right)}$ is given and fixed. We simply approximate the partial derivative of the first expectation with respect to $\Theta$ by computing that of \eqref{equation_A_2_4}.

For the update of $\Upsilon^{\left(t+1\right)}$, however, $\mathbb{E}_{q\left(\bg_X\middle|\bL^{\left(t\right)},\Upsilon^{\left(t\right)},\hat{\Gamma}\right)}\left[\log{p\left(\bg_X\middle|\Upsilon\right)}\right]$ can be expressed in a closed form, because both $p\left(\bg_X\middle|\Upsilon\right)$ and $q\left(\bg_X\middle| \bL^{\left(t\right)},\Upsilon^{\left(t\right)},\hat{\Gamma}^{\left(t\right)}\right)$ are Gaussian. Therefore, we have its exact partial derivative with respect to $\Upsilon$, which allows to use a gradient ascent method without sampling.

Unlike $\Theta^{\left(t+1\right)}$ and $\Upsilon^{\left(t+1\right)}$, we have a closed-form update of $\bL^{\left(t+1\right)}$ if $\pi \equiv 1$ in \eqref{equation_2_3_1}. By iteratively using the following identity \citep{Hutchinson1989}:
\begin{align*}
    \mathbb{E}\left[\left(Av+b\right)^\top{W}\left(Cv+d\right)\right] &=\left(Am+b\right)^\top 
    {W}\left(Cm+d\right)+\text{Trace}\left({\rm{S}}A^\top 
    {W}C\right),
\end{align*}
where ${W}$ is any $n\times n$ matrix, $b$ and $d$ $n$-dimensional vectors, $A$ and $C$ are $n\times k$ matrices, and $v$  is a $k$-dim random vector with mean $m$ and covariance matrix $S$, we can show the following:
\begin{equation}
\begin{aligned}
& \mathbb{E}_{q\left(\bbf_X,\bg_X\middle|\bL^{\left(t\right)},\Upsilon^{\left(t\right)},\hat{\Gamma}^{\left(t\right)}\right)}\left[\log{p\left(\bbf_X\middle| \bg_X;\bL\right)} + \log{p\left(\bL\middle|X\right)} \right] \\ = & -\frac{1}{2}\sum_{n=1}^{N}trace\left(\left(\sum_{d=1}^{D}{\omega_{\bx_nd}\blambda_d^{-1}}\right)\left(\mathbb{A}_{\bx_n}+\mathbb{B}_{\bx_n}\Omega_X\mathbb{B}_{\bx_n}^\mathbb{T}\right)\right)  - \frac{1}{2}\sum_{n=1}^{N}{\sum_{d=1}^{D}{\left({\omega_{\bx_nd}\cdot\log{\left|\blambda_d\right|}}\right)}} + \log{\pi\left(\bL\right)} + Const.,
\end{aligned}
\label{equation_A_2_6}
\end{equation}
where $\mathbb{A}$, $\mathbb{B}$ and $\Omega$ are defined as follows, for ${\hat{\Gamma}}^{\left(t\right)}=\left({\hat{\eta}}^{\left(t\right)},{\hat{\Psi}}^{\left(t\right)}\right)$:
\begin{equation}
\begin{aligned}
\mathbb{A}_{\bx_n} & \triangleq \left(\mathbb{I}\otimes\mathcal{J}_N^n\right)\mathbb{V}_{XX}^{\left(t\right)}\left(\mathbb{V}_{XX}^{\left(t\right)}+\Lambda_{XX}^{\left(t\right)}\right)^{-1}\Lambda_{XX}^{\left(t\right)}\left(\mathbb{I}\otimes\mathcal{J}_N^n\right)^\top
\\ \mathbb{B}_{\bx_n} & \triangleq \left(\mathbb{I}\otimes\mathcal{J}_N^n\right)\Lambda_{XX}^{\left(t\right)}\left(\mathbb{V}_{XX}^{\left(t\right)}+\Lambda_{XX}^{\left(t\right)}\right)^{-1} \\ \Omega_{X} & \triangleq \vvec\left({\hat{\eta}}_{X}^{\left(t\right)}-\hat{\mu}_{X}^{\left(t\right)}\right){\vvec\left({\hat{\eta}}_{X}^{\left(t\right)}-\hat{\mu}_{X}^{\left(t\right)}\right)}^\top+{\hat{\Psi}}_{XX}^{\left(t\right)},
\end{aligned}
\label{equation_A_2_9}
\end{equation}
where $\hat{\eta}_{X}^{\left(t\right)}$ is a $N \times Q$ matrix with each row being the transpose of $\hat{\eta}_n^{\left(t\right)}$ and $\hat{\Psi}_{XX}^{\left(t\right)} \triangleq \sum_{n=1}^{N}{({\hat{\Psi}}_n^{\left(t\right)}\otimes\mathcal{J}_N^{nn})}$.

Therefore, if $\pi \equiv 1$, the following equation,
\begin{equation*}
\begin{aligned}
\frac{\partial}{\partial\lambda_d}\mathbb{E}_{q\left(\bbf_X,\bg_X\middle|\bL^{\left(t\right)},\Upsilon^{\left(t\right)},\hat{\Gamma}^{\left(t\right)}\right)}\left[\log{p\left(\bbf_X\middle| \bg_X;\bL\right)} + \log{p\left(\bL\middle|X\right)} \right] = 0,
\end{aligned}
\label{equation_A_2_7}
\end{equation*}
implies that, for each $d$:
\begin{equation*}
\blambda_d=\frac{\sum_{n=1}^{N}{\omega_{\bx_n d}\left(\mathbb{A}_{\bx_n}+\mathbb{B}_{\bx_n} \Omega_{X} \mathbb{B}_{\bx_n}^\top\right)}}{\sum_{n=1}^{N}{\omega_{\bx_n d}}},
\label{equation_A_2_8}
\end{equation*}

Thus, we have the following {\it{closed-form}} update for $\bL^{\left(t+1\right)}$ if $\pi \equiv 1$:
\begin{equation}
\begin{aligned}
\bL^{\left(t+1\right)} =\left\{\blambda_d^{\left(t+1\right)}\right\}_{d=1}^{D}=\argmax_{\bL}{\tilde{\mathcal{Q}}\left(\bL,\Theta,\Upsilon\middle|\bL^{\left(t\right)},{\Theta}^{\left(t\right)},{\Upsilon}^{\left(t\right)}\right)} = \left\{\frac{\sum_{n=1}^{N}{\omega_{\bx_n d}\left(\mathbb{A}_{\bx_n}+\mathbb{B}_{\bx_n}\Omega_{X}{\mathbb{B}_{\bx_n}^\top}\right)}}{\sum_{n=1}^{N}{\omega_{\bx_n d}}}\right\}_{d=1}^{D}.
\end{aligned}
\label{equation_A_2_11}
\end{equation}

Note that $D=1$ results in the homoscedastic model, and the corresponding M-step update for $\bL^{\left(t+1\right)} = \left\{\blambda^{\left(t+1\right)}\right\}$ is nothing but:
\begin{equation*}
\blambda^{\left(t+1\right)} = \frac{1}{N}\sum_{n=1}^{N}{\left(\mathbb{A}_{\bx_n}+\mathbb{B}_{\bx_n}\Omega_{X}{\mathbb{B}_{\bx_n}^\top}\right)},
\label{equation_A_2_12}
\end{equation*}
because $\omega_{\bx_n d} \equiv 1$ for each $n$.

\section{A Practical Alternative to Update Kernel Hyperparameters}\label{secB}
In Sections \ref{secA_1} and \ref{secA_2}, we have discussed the standard setting of a VEM algorithm. However, the objective function in \eqref{equation_A_2_2} for updating $\Upsilon^{\left(t+1\right)}$ is involved with a numerical issue. Note that, for each kernel hyperparameter $\upsilon \in \Upsilon$:
\begin{equation}
\begin{aligned}
\frac{\partial}{\partial\upsilon}\mathbb{E}_{q\left(\bbf_X,\bg_X\middle|\bL^{\left(t\right)},\Upsilon^{\left(t\right)},\hat{\Gamma}\right)}\left[\log{p\left(\bg_X\middle|\Upsilon\right)}\right] = \frac{1}{2}\cdot trace\left(\mathbb{V}_{XX}^{-1}\left({\mathbb{C}}_X - \mathbb{V}_{XX}\right)\mathbb{V}_{XX}^{-1}\frac{\partial\mathbb{V}_{XX}}{\partial\upsilon}\right),
\end{aligned}
\label{equation_A_3_1}
\end{equation}
where ${\mathbb{C}}_X$ is defined as follows, for $\Omega_X$ in \eqref{equation_A_2_9}:
\begin{equation*}
\begin{aligned}
{\mathbb{C}}_X \triangleq \mathbb{V}_{XX}^{\left(t\right)}-\mathbb{V}_{XX}^{\left(t\right)}\left(\Lambda_{XX}^{\left(t\right)}+\mathbb{V}_{XX}^{\left(t\right)}\right)^{-1}\mathbb{V}_{XX}^{\left(t\right)} + \mathbb{V}_{XX}^{\left(t\right)}\left(\Lambda_{XX}^{\left(t\right)}+\mathbb{V}_{XX}^{\left(t\right)}\right)^{-1}\Omega_X\left(\Lambda_{XX}^{\left(t\right)}+\mathbb{V}_{XX}^{\left(t\right)}\right)^{-1}\mathbb{V}_{XX}^{\left(t\right)}.
\end{aligned}
\label{equation_A_3_2}
\end{equation*}

Because the partial derivative \eqref{equation_A_3_1} is involved with $\mathbb{V}_{XX}^{-1}$, our standard VEM cannot deal with the case that $\mathbb{V}_{XX}$ is degenerate as long as it resorts to a gradient ascent method: it could happen if two coordinates of responses are of true correlation 1 or -1, or some of the training covariates in $X$ are identical.

To deal with it, we can instead estimate the kernel hyperparameters $\Upsilon$ in the E-step when the variational parameters in $\Gamma$ are estimated to maximize the ELBO of $\log{p\left(Y\middle|X;\bL^{\left(t\right)},\Theta^{\left(t\right)},\Upsilon^{\left(t\right)}\right)}$ in \eqref{equation_A_1_2}, i.e., we restate it by following:
\begin{equation*}
\mathcal{L}\left(\Upsilon,\Gamma\right)  \triangleq {\mathbb{E}_{q\left(\bbf_X\middle|\Gamma\right)}\left[\log{\frac{p\left(Y\middle| \bbf_X;\Theta^{\left(t\right)}\right)p\left(\bbf_X\middle|\bL^{\left(t\right)},\Upsilon\right)}{q\left(\bbf_X\middle|\Gamma\right)}}\right]},
\label{equation_A_3_3}
\end{equation*}
and estimate not only $\Gamma$ but also $\Upsilon$ that maximize $\mathcal{L}\left(\Upsilon,\Gamma\right)$. Note that, for each kernel hyperparameter $\upsilon \in \Upsilon$:
\begin{equation}
\begin{aligned}
& \frac{\partial}{\partial\upsilon}\mathbb{E}_{q\left(\bbf_X\middle|\Gamma\right)}\left[\log{p\left(\bbf_X\middle|\bL^{\left(t\right)},\Upsilon\right)}\right] \\ & = \frac{1}{2}\cdot trace\left(\left(\mathbb{V}_{XX}+\Lambda_{XX}^{\left(t\right)}\right)^{-1}\left(\Omega_X - \mathbb{V}_{XX}-\Lambda_{XX}^{\left(t\right)} \right)\left(\mathbb{V}_{XX}+\Lambda_{XX}^{\left(t\right)}\right)^{-1}\frac{\partial\mathbb{V}_{XX}}{\partial\upsilon}\right),
\label{equation_A_3_4}
\end{aligned}
\end{equation}
which relies on $(\mathbb{V}_{XX}+\Lambda_{XX}^{\left(t\right)})^{-1}$ instead of $\mathbb{V}_{XX}^{-1}$. Note that $\mathbb{V}_{XX}+\Lambda_{XX}^{\left(t\right)}$ is invertible if any $\blambda_d \in \bL$ is positive-definite.

One interesting point is that \textit{both} of the fixed points of $\bbV_{XX}$, $\Omega_X$ and $\Lambda_{XX}$ of \eqref{equation_A_3_1} and those of \eqref{equation_A_3_4} satisfy the equality $\Omega_X = \bbV_{XX} + \Lambda_{XX}$. The underlying intuition is that $\Omega_X$ is the empirical covariance of $\vvec(\bbf_X)$ from the variational parameters $\Gamma=(\eta,\Psi)$ and $\bbV_{XX}+\Lambda_{XX}$ is the theoretical covariance of $\vvec(\bbf_X)$ from the modeling.

\section{Bandwidth determination for the precision process}
\label{secB1}

Here we describe how we have chosen the bandwidths $\mathcal{H}=\left\{h_d\right\}_{d=1}^{D}$ of the density kernel $\mathcal{K}$ to define the weight vector in \eqref{equation_2_1_2}, which is essential in defining Algorithm \ref{algorithm1} of Section \ref{sec3_6}. Following the idea of \cite{LANGRENE2019}, we choose each $h_d$ as a function of the percentage $r$ and the feature data $X=\left\{\bx_n\right\}_{n=1}^{N}$ to be covered. To be specific, each $h_d$ is defined by the average of the minimum and maximum diameters such that the corresponding neighborhood of the induced point ${\underline{\bx}}_d$ contains exactly $\left\lceil r\cdot{N}/100\right\rceil$ elements of $X$. This ``$r$\%-nearest-neighbors'' rule is adaptive to the disparity between the distribution of $X$ and that of the induced points $\underline{X}=\left\{{\underline{\bx}}_d\right\}_{d=1}^{D}$.

To choose an appropriate  value of $r\in(0,100)$, we first pick and fix a hyperparameter $A\in(0,100)$, which we call the``adjacent percentage parameter'', and  prepare a set $\mathcal{R}$ of candidate $r$'s. Then, we conduct the following cross-validation procedure after each iteration of the variational EM algorithm, assuming that $\mathbb{A}$, $\mathbb{B}$ and $\Omega$ in \eqref{equation_A_2_9} are fixed constants from $\Theta=\Theta^{(t+1)}$, $\Upsilon=\Upsilon^{(t+1)}$ and $\Gamma=\hat{\Gamma}$ updated in the previous M-step:
\begin{enumerate}
\item For each $\bx_n\in{X}$, define a set ${\underline{X}}^{\left(n\right)}\subseteq\underline{X}$ that contains exactly $\left\lceil{AD}/100\right\rceil$ inducing points $\underline{\bx}_d\in \underline{X}$ closest to $\bx_n$.
\item For each candidate percentage $r\in\mathcal{R}$, define the corresponding $\mathcal{H}^{(r)}=\left\{h_d^{(r)}\right\}_{d=1}^{D}$, and do the following:
\begin{enumerate}
\item Compute $\hat{\bL}^{(r)}=\left\{\hat{\blambda}_{d}^{(r)}\right\}_{d=1}^D$ by \eqref{equation_A_2_11}, with $\omega_{\bx_n d}$'s being defined as \eqref{equation_2_1_2} for the bandwidth $\mathcal{H}^{(r)}$.
\item Compute $\tilde{\omega}_{\bx_n}^{(r)} = \left\{\tilde{\omega}_{\bx_n d}^{(r)}\right\}_{d=1}^D$ for each $n$ as follows:
\begin{equation*}
\tilde{\omega}_{\bx_nd}^{\left(r\right)} \triangleq \frac{1_{\left\{{\underline{\bx}}_d\notin{\underline{X}}^{\left(n\right)}\right\}}\cdot\mathcal{K}_{h_d^{\left(r\right)}}\left(\bx_n,{\underline{\bx}}_d\right)}{\sum_{{\underline{\bx}}_c\notin{\underline{X}}^{\left(n\right)}}{\mathcal{K}_{h_c^{\left(r\right)}}\left(\bx_n,{\underline{\bx}}_c\right)}}.
\label{equation_3_5_0}
\end{equation*}
\item Compute the following for $\hat{\bL}^{(r)}$ and ${\widetilde{\omega}}^{\left(r\right)}=\left\{{\widetilde{\omega}}_{\bx_n}^{\left(r\right)}\right\}_{n=1}^N$:
\begin{equation}
\begin{aligned}
\mathcal{T}\left(r\right) \triangleq & \ \mathbb{E}_{q\left(\bbf_X,\bg_X\middle|\bL^{\left(t\right)},\Theta^{\left(t\right)},\hat{\Gamma}^{\left(t\right)}\right)}\left[\log{p\left(\bbf_X\middle| \bg_X;\hat{\bL}^{(r)},{\widetilde{\omega}}^{\left(r\right)}\right)}\right] \\ = & -\frac{1}{2}\sum_{n=1}^{N}\text{Trace} \left(\left({\widetilde{\Lambda}}^{\left(r\right)}\left(\bx_n\right)\right)^{-1}\left(\mathbb{A}_{\bx_n}+\mathbb{B}_{\bx_n}\Omega_X\mathbb{B}_{\bx_n}^\mathbb{T}\right)\right) \\ & - \frac{1}{2}\sum_{n=1}^{N}\log{\left|{\widetilde{\Lambda}}^{\left(r\right)}\left(\bx_n\right)\right|}-\frac{1}{2}NQ\log{\left|2\pi\right|},
\end{aligned}
\label{equation_3_5_1}
\end{equation}
where ${\widetilde{\Lambda}}^{\left(r\right)}\left(\bx_n\right) \triangleq \left(\sum_{d=1}^{D}{{\widetilde{\omega}}_{\bx_{n}d}^{\left(r\right)}\left({\hat{\blambda}}_d^{\left(r\right)}\right)^{-1}}\right)^{-1}$ for each $n$.
\end{enumerate}
\item Pick $\hat{r}\triangleq \argmax_{r \in \mathcal{R}}{\mathcal{T}\left(r\right)}$.
\end{enumerate}

In words, each $\tilde{\omega}_{\bx_nd}^{(r)}$ computed 
for the cross-validation is proportional to $\mathcal{K}_{h_d^{\left(r\right)}}\left(\bx_n,{\underline{\bx}}_d\right)$ as usual, except for some nearest ${\underline{\bx}}_d$'s of $\bx_n$ that result in $\tilde{\omega}_{\bx_nd}^{(r)}=0$. Also, each ${\widetilde{\Lambda}}^{\left(r\right)}\left(\bx_n\right)=\left(\sum_{d=1}^{D}{{\widetilde{\omega}}_{\bx_nd}^{\left(r\right)}\left({\hat{\blambda}}_d^{\left(r\right)}\right)^{-1}}\right)^{-1}$ means that $\hat{\blambda}_d^{(r)}$'s at some nearest ${\underline{\bx}}_d$'s of $\bx_n$ are not supposed to contribute in the mixture of precision. The expectation \eqref{equation_3_5_1} is a modified version of the objective function for updating $\bL$ defined in \eqref{equation_3_1_11} and \eqref{equation_A_2_6}, by removing the prior term $\log{p( \bL \mid X )}$.

Note that $\mathcal{T}\left(r\right)$ can be computed \textit{coordinate-wisely} if each $\blambda_d$ is supposed to be a diagonal covariance matrix, which makes the generalization in Appendix \ref{secF} tractable for a large $Q$. We have tested this cross-validation with both simulated and real data in Section \ref{sec4_0} and \ref{sec5}, to see whether $r$ converges over iterations.

\section{Interpretation of the VEM-based Inference}\label{secB2}

Our variational EM (VEM) algorithm depends on the mean-field approximation to the posterior of the hidden variables to define the Q-function. For example, HeGPR-H in Section \ref{sec4_1_3} approximates the posterior distribution $p\left(\bbf_X,\bg_X,\balpha\middle|Y,X;\bP^{\left(t\right)},\bL^{\left(t\right)},\Upsilon^{\left(t\right)}\right)$ with $q\left(\bbf_X,\bg_X,\balpha\middle|\bL^{\left(t\right)},\Upsilon^{\left(t\right)},\Gamma\right)$ defined in \eqref{equation_4_1_3_6} and \eqref{equation_4_1_3_7}, and then approximates the Q-function as follows:
\begin{equation}
Q\left(\bP,\bL,\Theta,\Upsilon\middle|\bP^{\left(t\right)},\bL^{\left(t\right)},\Theta^{\left(t\right)},\Upsilon^{\left(t\right)}\right)\approx\mathbb{E}_{q\left(\bbf_X,\bg_X,\balpha\middle|\bL^{\left(t\right)},\Upsilon^{\left(t\right)},\Gamma\right)}\left[\log{p\left(Y,\bbf_X,\bg_X,\bL,\balpha,\bP\middle|X,\Upsilon\right)}\right].
\label{equation_B2_1}
\end{equation}

Because $q\left(\bbf_X,\bg_X,\balpha\middle|\bL^{\left(t\right)},\Upsilon^{\left(t\right)},\Gamma\right)=q\left(\bbf_X\middle|\Gamma\right)q\left(\balpha\middle|\Gamma\right)p\left(\bg_X\middle| \bbf_X,\bL^{\left(t\right)},\Upsilon^{\left(t\right)}\right)$, the above \eqref{equation_B2_1} is nothing but:
\begin{equation*}
\begin{aligned}
& \mathbb{E}_{q\left(\bbf_X,\bg_X,\balpha\middle|\bL^{\left(t\right)},\Upsilon^{\left(t\right)},\Gamma\right)}\left[\log{p\left(Y,\bbf_X,\bg_X,\bL,\balpha,\bP\middle|X,\Upsilon\right)}\right] \\ & =\mathbb{E}_{q\left(\bbf_X\middle|\Gamma\right)p\left(\bg_X\middle| \bbf_X,\bL^{\left(t\right)},\Upsilon^{\left(t\right)}\right)}\left[\mathbb{E}_{q\left(\balpha\middle|\Gamma\right)}\left[\log{p\left(Y,\bbf_X,\bg_X,\bL,\balpha,\bP\middle|X,\Upsilon\right)}\right]\right],
\label{equation_B2_2}
\end{aligned}
\end{equation*}
and:
\begin{equation*}
\begin{aligned}
& \mathbb{E}_{q\left(\balpha\middle|\Gamma\right)}\left[\log{p\left(Y,\bbf_X,\bg_X,\bL,\balpha,\bP\middle|X,\Upsilon\right)}\right] \\ & =\mathbb{E}_{q\left(\balpha\middle|\Gamma\right)}\left[\log{p\left(Y\middle| \bbf_X,\balpha,\bP\right)}\right]+\log{p\left(\bbf_X\middle| \bg_X;\bL\right)} +\log{p\left(\bg_X\middle|\Upsilon\right)} +\mathbb{E}_{q\left(\balpha\middle|\Gamma\right)}\left[\log{p\left(\balpha\right)}\right]+\log{p\left(\bL\middle|X\right)}+\log{p\left(\bP\middle|X\right)}.
\label{equation_B2_3}
\end{aligned}
\end{equation*}

Note that, for constants $\mathcal{C}$ and $\widetilde{\mathcal{C}}$, we have:
\begin{equation*}
\begin{aligned}
& \mathbb{E}_{q\left(\balpha\middle|\Gamma\right)}\left[\log{p\left(Y\middle| \bbf_X,\balpha,\bP\right)}\right]+\log{p\left(\bbf_X\middle| \bg_X;\bL\right)} \\ = & -\frac{1}{2}\sum_{n=1}^{N}{\xi_n^2\left(\by_n-\bbf\left(\bx_n\right)\right)^\top{\Phi\left(\bx_n\right)}^{-1}\left(\by_n-\bbf\left(\bx_n\right)\right)}-\frac{1}{2}\sum_{n=1}^{N}\log{\left|\xi_n^{-2}\Phi\left(\bx_n\right)\right|} \\ & -\frac{1}{2}\sum_{n=1}^{N}{\left(\bbf\left(\bx_n\right)-\bg\left(\bx_n\right)\right)^\top {\Lambda\left(\bx_n\right)}^{-1}\left(\bbf\left(\bx_n\right)-\bg\left(\bx_n\right)\right)}-\frac{1}{2}\sum_{n=1}^{N}\log{\left|\Lambda\left(\bx_n\right)\right|}+\mathcal{C} \\ 
= & \sum_{n=1}^{N}\log{\mathcal{N}\left(\by_n\middle| \bbf\left(\bx_n\right),\xi_n^{-2}\Phi\left(\bx_n\right)\right)}+\sum_{n=1}^{N}\log{\mathcal{N}\left(\bbf\left(\bx_n\right)\middle| \bg\left(\bx_n\right),\Lambda\left(\bx_n\right)\right)}+\widetilde{\mathcal{C}},
\label{equation_B2_4}
\end{aligned}
\end{equation*}
thus, our VEM algorithm \textit{implicitly} assumes the two-layer Gaussian model on $\by_n$'s \textit{given} $\xi_n$’s and $\bg$. In other words, the original model $\by_n \sim \mathcal{T}_\nu\left(\bbf\left(\bx_n\right),\Phi\left(\bx_n\right)\right)$ in \eqref{equation_4_1_3_3} is interpreted as $\by_n \sim \mathcal{N}\left(\bbf\left(\bx_n\right),\xi_n^{-2}\Phi\left(\bx_n\right)\right)$ \textit{given} $\xi_n$’s at the inference stage.

\section{A Bayesian Approach Toward the Estimation}\label{secC}
Our standard VEM setting does not consider priors on $\Theta$ and $\Upsilon$, and $\pi \equiv 1$ for $\bL$. The model is easily extendable to be Bayesian with priors $p_0(\Theta)$ and $p_0(\Upsilon)$: the ``complete-data'' likelihood becomes:
\begin{equation*}
\begin{aligned}
p\left(Y,\bbf_X,\bg_X,\bL,\Theta,\Upsilon\right) = p\left(Y\middle| \bbf_X;\Theta\right)p\left(\bbf_X\middle| \bg_X;\bL\right)p\left(\bg_X\middle|\Upsilon\right) p\left(\bL\middle|X\right) \cdot p_0(\Theta) p_0(\Upsilon),
\end{aligned}
\label{equation_A_3_5}
\end{equation*}
and the objective functions for updating $\Theta = \Theta^{\left(t+1\right)}$ and $\Upsilon = \Upsilon^{\left(t+1\right)}$ are:
\begin{equation*}
\begin{aligned}
\Theta^{\left(t+1\right)} & \triangleq \argmax_{\Theta}{\mathbb{E}_{q\left(\bbf_X\middle|\hat{\Gamma}^{\left(t\right)}\right)}\left[\log{p\left(Y\middle| \bbf_X;\Theta\right)} + \log{p_0(\Theta)} \right]} \\ \Upsilon^{\left(t+1\right)} & \triangleq \argmax_{\Upsilon}{\mathbb{E}_{q\left(\bff_X,\bg_X\middle| \bL^{\left(t\right)},{\Upsilon}^{\left(t\right)},\hat{\Gamma}^{\left(t\right)}\right)}\left[\log{p\left(\bg_X\middle|\Upsilon\right)} + \log{p_0(\Upsilon)}  \right]}.
\end{aligned}
\label{equation_A_3_7}
\end{equation*}

In particular, if $\pi$ in \eqref{equation_2_3_1} is the inverse-Wishart prior for a fixed pair of hyperparameters $\left(\Gamma_0,\nu_0\right)$:
\begin{equation*}
\pi\left(\blambda_d\right) = \mathcal{W}^{-1}\left(\blambda_d\middle|\Gamma_0,\nu_0\right),
\label{equation_2_2_8}
\end{equation*}
then we still have a closed-form update of $\bL^{\left(t+1\right)}$ in the M-step as follows:
\begin{equation*}
\begin{aligned}
\bL^{\left(t+1\right)} = \left\{\frac{\Gamma_0 + \sum_{n=1}^{N}{\omega_{\bx_n d}\left(\mathbb{A}_{\bx_n}+\mathbb{B}_{\bx_n}\Omega_{X}{\mathbb{B}_{\bx_n}^\top}\right)}}{\nu_0 + {Q} + 1 + \sum_{n=1}^{N}{\omega_{\bx_n d}}}\right\}_{d=1}^{D}.
\end{aligned}
\label{equation_A_3_6}
\end{equation*}

\section{An Alternative Method Based on Sampling}\label{secD}

In Section \ref{sec3_1}, we mentioned an alternative approach toward the posterior approximation based on the posterior samples. Suppose that we have drawn a set of $M$ i.i.d. samples ${\widetilde{\bbf}}_X=\left\{{\widetilde{\bbf}}_X^{\left(m\right)}\right\}_{m=1}^M$ from the marginal posterior, $p\left(\bbf_X\middle|Y,X;\bL^{\left(t\right)},\Theta^{\left(t\right)},\Upsilon^{\left(t\right)}\right)\propto p\left(Y\middle| \bbf_X;\Theta^{\left(t\right)}\right)p\left(\bbf_X\middle|\bL^{\left(t\right)},\Upsilon^{\left(t\right)}\right)$, where:
\begin{equation*}
\begin{aligned}
p\left(\bbf_X\middle|\bL^{\left(t\right)},\Upsilon^{\left(t\right)}\right) \triangleq \mathbb{E}_{p\left(\bg_X\middle|\Upsilon^{\left(t\right)}\right)}\left[p\left(\bbf_X\middle| \bg_X;\bL^{\left(t\right)}\right)\right] =\mathcal{N}\left(\vvec\left(\bbf_X\right)\middle| \vvec \left(\mu_X^{\left(t\right)}\right),\mathbb{V}_{XX}^{\left(t\right)}+\Lambda_{XX}^{\left(t\right)}\right).
\end{aligned}
\label{equation_D_1}
\end{equation*}
Then, the posterior \eqref{equation_3_1_5} is approximated as follows:
\begin{equation*}
\begin{aligned}
p\left(\bbf_X,\bg_X\middle|Y,X;\bL^{\left(t\right)},\Theta^{\left(t\right)},\Upsilon^{\left(t\right)}\right) \approx\frac{1}{M}\sum_{m=1}^{M}{1_{\left\{\bbf_X={\widetilde{\bbf}}_X^{\left(m\right)}\right\}}\left(\bbf_X\right) \cdot p\left(\bg_X\middle| \bbf_X;\bL,\Upsilon\right)},
\end{aligned}
\label{equation_D_2}
\end{equation*}
which results in the following approximation of the Q-function \eqref{equation_3_1_3}:
\begin{equation*}
\begin{aligned}
\mathcal{Q}\left(\bL,\Theta,\Upsilon\middle|\bL^{\left(t\right)},{\Theta}^{\left(t\right)},{\Upsilon}^{\left(t\right)}\right) \approx \frac{1}{M}\sum_{m=1}^{M}{\mathbb{E}_{p\left(\bg_X\middle|{\widetilde{\bbf}}_X^{\left(m\right)};\bL^{\left(t\right)},\Upsilon^{\left(t\right)}\right)}\left[\log{p\left(Y,{\widetilde{\bbf}}_X^{\left(m\right)},\bg_X,\bL\middle|X;\Theta,\Upsilon\right)}\right]}.
\end{aligned}
\label{equation_D_3}
\end{equation*}

Therefore, the M-step updates are given as follows:
\begin{equation*}
\begin{aligned}
{\Theta}^{\left(t+1\right)} & \triangleq \argmax_{\Theta}{\sum_{m=1}^{M}\log{p\left(Y\middle|{\widetilde{\bbf}}_X^{\left(m\right)};\Theta\right)}} \\ {\Upsilon}^{\left(t+1\right)} & \triangleq \argmax_{\Upsilon}{\sum_{m=1}^{M}{\mathbb{E}_{p\left(\bg_X\middle|{\widetilde{\bbf}}_X^{\left(m\right)};\bL^{\left(t\right)},\Upsilon^{\left(t\right)}\right)}\left[\log{p\left(\bg_X\middle|\Upsilon\right)}\right]}} \\ {\bL}^{\left(t+1\right)} & \triangleq \argmax_{\bL}{\sum_{m=1}^{M}{\mathbb{E}_{p\left(\bg_X\middle|{\widetilde{\bbf}}_X^{\left(m\right)};\bL^{\left(t\right)},\Upsilon^{\left(t\right)}\right)}\left[\log{p\left({\widetilde{\bbf}}_X^{\left(m\right)}\middle| \bg_X;\bL\right)}+\log{p\left(\bL\middle|X\right)}\right]}}.
\end{aligned}
\label{equation_D_4}
\end{equation*}

Also, we have the following {\it{closed-form}} update for $\bL^{\left(t+1\right)}$ if $\pi \equiv 1$ in \eqref{equation_2_3_1}:
\begin{equation*}
\begin{aligned}
& \bL^{\left(t+1\right)} =\left\{\blambda_d^{\left(t+1\right)}\right\}_{d=1}^{D} = \left\{\frac{\sum_{n=1}^{N}{\omega_{\bx_n d}\left(\mathbb{A}_{\bx_n}+\mathbb{B}_{\bx_n}\Omega_{X}{\mathbb{B}_{\bx_n}^\top}\right)}}{\sum_{n=1}^{N}{\omega_{\bx_n d}}}\right\}_{d=1}^{D},
\end{aligned}
\label{equation_D_5}
\end{equation*}
where:
\begin{equation*}
\begin{aligned}
\mathbb{A}_{\bx_n} & \triangleq \left(\mathbb{I}\otimes\mathcal{J}_N^n\right)\mathbb{V}_{XX}^{\left(t\right)}\left(\mathbb{V}_{XX}^{\left(t\right)}+\Lambda_{XX}^{\left(t\right)}\right)^{-1}\Lambda_{XX}^{\left(t\right)}\left(\mathbb{I}\otimes\mathcal{J}_N^n\right)^\top
\\ \mathbb{B}_{\bx_n} & \triangleq \left(\mathbb{I}\otimes\mathcal{J}_N^n\right)\Lambda_{XX}^{\left(t\right)}\left(\mathbb{V}_{XX}^{\left(t\right)}+\Lambda_{XX}^{\left(t\right)}\right)^{-1} \\ \Omega_{X} & \triangleq \frac{1}{M}\sum_{m=1}^{M}{\vvec\left({\widetilde{\bbf}}_X^{\left(m\right)}-\mu_X^{\left(t\right)}\right){\vvec\left({\widetilde{\bbf}}_X^{\left(m\right)}-\mu_X^{\left(t\right)}\right)}^\top}.
\end{aligned}
\label{equation_D_6}
\end{equation*}

\section{Handling Missing Response Values}\label{secE}

So far, we have discussed the modeling when each $Q$-dimensional response $y_n$ is fully observed, i.e., there is no missing value in $\left\{y_{n1},y_{n2},\cdots,y_{n{Q}}\right\}$ for each $n$. Thus, we can call the previous setting  a multivariate Gaussian process model [\cite{Rakitsch2013}].

Let us consider a more general case in which each (multivariate) response $y_n$ may have missing values at its certain components. This is more likely a case that we have multiple datasets from different data sources over the same covariates but expect that they are correlated, so it is natural to expect that the responses of datasets are not synchronized one another.

To deal with it, for:
\begin{equation*}
\begin{aligned}
{Y}^{\left(0\right)} \triangleq\left\{y_{nq}\middle| y_{nq}\ is\ missing\right\}={Y} \backslash {Y}^{\left(1\right)}  \ \ \mbox{and} \ \  \bff_X^{\left(0\right)}  \triangleq\left\{f_{nq}\middle| y_{nq}\ is\ missing\right\}=\bff_X \backslash \bff_X^{\left(1\right)},
\end{aligned}
\label{equation_E_1}
\end{equation*}
we have the following ELBO $\mathcal{L}\left(\Gamma\right)$ of $\log{p\left(Y^{\left(1\right)}\middle|X,\bL^{\left(t\right)},\Theta^{\left(t\right)},\Upsilon^{\left(t\right)}\right)}$ from \eqref{equation_A_1_2}:
\begin{equation*}
\begin{aligned}
\mathcal{L}\left(\Gamma\right) \triangleq \mathbb{E}_{q\left(\bbf_X\middle|\Gamma\right)}\left[\log{p\left(\bbf_X\middle|\bL^{\left(t\right)},\Upsilon^{\left(t\right)}\right)} - \log{q\left(\bbf_X\middle|\Gamma\right)}\right] + \mathbb{E}_{q\left(\bbf_X^{\left(1\right)}\middle|\Gamma\right)}\left[\log{p\left(Y^{\left(1\right)}\middle| \bbf_X^{\left(1\right)};\Theta^{\left(t\right)}\right)}\right]
\end{aligned}
\label{equation_E_2}
\end{equation*}

That is, we estimate parameters as if there were no third-level model for missing $Y^{(0)}$. Variational parameters that are coupled with $Y^{(1)}$ are estimated from both observed $Y^{(1)}$ and its prior plus entropy, while those coupled with $Y^{(0)}$ are estimated only from its prior plus entropy. The marginal model $p\left(\bbf_X\middle|\bL^{\left(t\right)},\Upsilon^{\left(t\right)}\right) = \mathcal{N}\left(\vvec{\left(\bbf_X\right)}\middle| \vvec{\left(\mu_X\right)},\mathbb{V}_{XX}^{\left(t\right)}+\Lambda_X^{\left(t\right)}\right)$ \textit{indirectly} transfers the information from $\bbf_X^{(1)}$ to $\bbf_X^{(0)}$ based on their correlation.

The update in the M-step is given as follows:
\begin{equation*}
\begin{aligned}
\Theta^{\left(t+1\right)} & \triangleq \argmax_{\Theta}{\mathbb{E}_{q\left(\bbf_X^{\left(1\right)}\middle|\hat{\Gamma}^{\left(t\right)}\right)}\left[\log{p\left(Y^{\left(1\right)}\middle| \bbf_X^{\left(1\right)};\Theta\right)}\right]} \\ \Upsilon^{\left(t+1\right)} & \triangleq \argmax_{\Upsilon}{\mathbb{E}_{q\left(\bbf_X,\bg_X\middle|L^{\left(t\right)},\Upsilon^{\left(t\right)},\hat{\Gamma}^{\left(t\right)}\right)}\left[\log{p\left(\bg_X\middle|\Upsilon\right)}\right]} \\ \bL^{\left(t+1\right)} & \triangleq \argmax_{\bL}{\mathbb{E}_{q\left(\bbf_X,\bg_X\middle|\bL^{\left(t\right)},\Theta^{\left(t\right)},\hat{\Gamma}^{\left(t\right)}\right)}\left[\log{p\left(\bbf_X\middle| \bg_X,\bL\right)} + \log{p\left(\bL\middle|X\right)} \right]}.
\end{aligned}
\label{equation_E_3}
\end{equation*}

\section{Handling Heterogeneity between Response Components}\label{secF}

So far, we have assumed that the conditional model of $\bbf(\bx)$ given $\bg(\bx)$ is governed by a set $\bL$ of full-rank $Q \times Q$ matrices $\left\{\blambda_d\right\}_{d=1}^D$ and a \textit{single} set of bandwidths $\mathcal{H}$. This can be generalized into several ways, including the following modification of the second-level model \eqref{equation_2_1_1} in Section \ref{sec2_2}:
\begin{equation}
\left[\bbf\left(\bx\right)\right]_q \sim \mathcal{N}\left(\left[\bg\left(\bx\right)\right]_q,\Lambda_{q}\left(\bx\right)\right), \quad q = 1,\cdots,Q,
\label{equation_G_1}
\end{equation}
for $\Lambda_q\left(\bx\right) \triangleq \left(\sum_{d=1}^{D^{\left(q\right)}}{\omega_{\bx d}^{\left(q\right)}\blambda_{dq}^{-1}}\right)^{-1}$, so that:
\begin{equation}
p\left(\bbf\left(\bx\right)\middle| \bg\left(\bx\right);\bL\right) = \prod_{q=1}^{Q}{\mathcal{N}\left(\left[\bbf\left(\bx\right)\right]_q\middle|\left[\bg\left(\bx\right)\right]_q,\Lambda_q\left(\bx\right)\right)},
\label{equation_G_2}
\end{equation}
where $\left[\bbf\left(\bx\right)\right]_q$ and $\left[\bg\left(\bx\right)\right]_q$ are the $q$-th components of $\bbf\left(\bx\right)$ and $\bg\left(\bx\right)$, respectively, and $\omega_{\bx}^{\left(q\right)}\triangleq\left\{\omega_{\bx d}^{\left(q\right)}\right\}_{d=1}^{D^{\left(q\right)}}$ are component-wise weights for a set of bandwidth $\mathcal{H}^{(q)}$. Each $\blambda_{dq}$ is a component-wise positive variance parameter. 

Clearly, \eqref{equation_G_2} together with \eqref{equation_G_1} implies that heteroscedastic variances could show heterogeneity between response components. It is straightforward to obtain similar results described in Section \ref{sec3} and Appendix \ref{secA} with the following modification of $\Lambda$ in the second-level model \eqref{equation_2_1_1} and each $\blambda_{dq} \in \bL^{\left(q\right)} \in \bL$: $\Lambda\left(\bx\right) \triangleq diag\left(\Lambda_1\left(\bx\right),\cdots,\Lambda_Q\left(\bx\right)\right)$ is a diagonal matrix with its $q$-th diagonal entry being $\Lambda_q\left(\bx\right) = \left(\sum_{d=1}^{D^{\left(q\right)}}{\omega_{\bx d}^{\left(q\right)}\blambda_{dq}^{-1}}\right)^{-1}$, and we assume a new prior $p\left(\bL\middle|X\right) = \prod_{q=1}^{Q}{p\left(\bL^{\left(q\right)}\middle|X\right)}$, where
\begin{equation*}
\begin{aligned}
p\left(\bL^{\left(q\right)}\middle|X\right)  \propto\exp\left(\frac{1}{2}\sum_{n=1}^{N}\sum_{d=1}^{D^{\left(q\right)}}{\omega_{\bx_nd}^{\left(q\right)}\log{\left(\blambda_{dq}^{-1}\right)}}-\frac{1}{2}\sum_{n=1}^{N}\log{\left(\sum_{d=1}^{D^{\left(q\right)}}{\omega_{\bx_nd}^{\left(q\right)}\blambda_{dq}^{-1}}\right)}\right)\cdot \pi\left(\bL\right).
\end{aligned}
\label{equation_G_3}
\end{equation*}

\section{Variational Approximation for Large Datasets}\label{secG}

One of the major practical obstacles in applying Gaussian process models to real data is the intractability of matrix inversion when the size of matrix is large. To overcome this drawback, various approximation methods have been devised [\cite{Bauer2016}], among which we consider the variational free energy method [\cite{Titsias2009}]. The method starts by choosing a small set of {\it induced covariates}, $\underline{X}$, which has a much smaller size than $X$ and can be understood as ``sufficient representatives'' (e.g., a few cluster centers) of the observed features of $X$. Intuitively, if $\underline{X}$ are nicely chosen so that the observations are nearly independent of each other conditional on $\underline{\rX}$ and its associated random function values, then, for any 
$\bx,\bx'\notin \underline{\rX}$, we have:
\begin{equation*}
\mathbb{V}\left(\bx,\bx'\right)\approx\mathbb{V}_{\bx\underline{X}}\mathbb{V}_{\underline{X}\underline{X}}^{-1}\mathbb{V}_{\underline{X}\bx'}.
\label{equation_3_8_1}
\end{equation*}
Moreover, for all $i\neq j$
we can approximate the $(i,j)$-th entry of the ${NQ}\times {NQ}$ kernel matrix as:
\begin{equation*}
\left[\mathbb{V}_{XX}\right]_{ij}\approx\left[\mathbb{V}_{X\underline{X}}\mathbb{V}_{\underline{X}\underline{X}}^{-1}\mathbb{V}_{\underline{X}X}\right]_{ij},
\label{equation_3_8_2}
\end{equation*}
and note that the diagonal elements $\bbV(\bx_n,\bx_n)$ can still be retained.

With this approximation, one can rewrite the VEM algorithm described in the main body accordingly. Here we just state the results. In Section \ref{secA_1}, the term $\log{p\left(\bbf_X\middle|\bL^{\left(t\right)},\Upsilon^{\left(t\right)}\right)}$ in the evidence lower bound \eqref{equation_A_1_2} has the following lower bound [\cite{Titsias2009}]:
\begin{equation*}
\begin{aligned}
& \log{p\left(\bbf_X\middle|\bL^{\left(t\right)},\Upsilon^{\left(t\right)}\right)} =\log{\mathcal{N}\left(\vvec\left(\bbf_X\right)\middle| \vvec\left(\mu_X^{\left(t\right)}\right),\mathbb{V}_{XX}^{\left(t\right)}+\Lambda_{XX}^{\left(t\right)}\right)} \\ & \geq\log{\mathcal{N}\left(\vvec\left(\bbf_X\right)\middle| \vvec \left(\mu_X^{\left(t\right)}\right),\mathbb{V}_{X\underline{X}}^{\left(t\right)}\left(\mathbb{V}_{\underline{X}\underline{X}}^{\left(t\right)}\right)^{-1}\mathbb{V}_{\underline{X}X}^{\left(t\right)}+\Lambda_{XX}^{\left(t\right)}\right)} \\ & - \frac{1}{2}\cdot trace\left(\left(\Lambda_{XX}^{\left(t\right)}\right)^{-1}\left(\mathbb{V}_{XX}^{\left(t\right)}-\mathbb{V}_{X\underline{X}}^{\left(t\right)}\left(\mathbb{V}_{\underline{X}\underline{X}}^{\left(t\right)}\right)^{-1}\mathbb{V}_{\underline{X}X}^{\left(t\right)}\right)\right),
\end{aligned}
\label{equation_F_3}
\end{equation*}
which brings a new evidence lower bound of $\log{p\left(Y\middle|X,\bL^{\left(t\right)},\Theta^{\left(t\right)},\Upsilon^{\left(t\right)}\right)}$ as the objective function for estimating $\Gamma$ (and $\Upsilon$ if we accept the alternative in Appendix \ref{secB}).

In Section \ref{secA_2}, $\mathbb{A}$ and $\mathbb{B}$ in equation \eqref{equation_A_2_9} are redefined as follows:
\begin{equation*}
\begin{aligned}
\mathbb{A}_{\bx_n} & \triangleq \left(\mathbb{I}\otimes\mathcal{J}_N^n\right)\left(\mathbb{V}_{XX}^{\left(t\right)}+\Lambda_{XX}^{\left(t\right)}-\mathbb{V}_{X\underline{X}}^{\left(t\right)}\left(\mathbb{V}_{\underline{X}\underline{X}}^{\left(t\right)}\right)^{-1}\mathbb{V}_{\underline{X}X}^{\left(t\right)}\right)\left(\mathbb{I}\otimes\mathcal{J}_N^n\right)^\top \\ & -\left(\mathbb{I}\otimes\mathcal{J}_N^n\right)\left(\Lambda_{XX}^{\left(t\right)}\left(\Lambda_{XX}^{\left(t\right)}+\mathbb{V}_{X\underline{X}}^{\left(t\right)}\left(\mathbb{V}_{\underline{X}\underline{X}}^{\left(t\right)}\right)^{-1}\mathbb{V}_{\underline{X}X}^{\left(t\right)}\right)^{-1}\Lambda_{XX}^{\left(t\right)}\right)\left(\mathbb{I}\otimes\mathcal{J}_N^n\right)^\top
\\ \mathbb{B}_{\bx_n} & \triangleq \left(\mathbb{I}\otimes\mathcal{J}_N^n\right)\Lambda_{XX}^{\left(t\right)}\left(\Lambda_{XX}^{\left(t\right)}+\mathbb{V}_{X\underline{X}}^{\left(t\right)}\left(\mathbb{V}_{\underline{X}\underline{X}}^{\left(t\right)}\right)^{-1}\mathbb{V}_{\underline{X}X}^{\left(t\right)}\right)^{-1}.
\end{aligned}
\label{equation_F_4}
\end{equation*}

By replacing $\mathbb{A}$ and $\mathbb{B}$ in equation \eqref{equation_A_2_11} with those in the above reformulation, one can still update $\bL^{\left(t+1\right)}$ exactly in the M-step. 
Note that the covariance matrices $\Lambda_{XX}^{\left(t\right)}+\mathbb{V}_{X\underline{X}}^{\left(t\right)}\left(\mathbb{V}_{\underline{X}\underline{X}}^{\left(t\right)}\right)^{-1}\mathbb{V}_{\underline{X}X}^{\left(t\right)}$ and $\Lambda_{XX}^{\left(t\right)}$ can be inverted efficiently (i.e., its time complexity is proportional to the size of data $N$) by the Woodbury matrix identity [\cite{Max1950}] and the following formula:
\begin{equation*}
\Lambda_{XX}^{-1}=\left(\sum_{n=1}^{N}\left(\Lambda\left(\bx_n\right)\otimes\mathcal{J}_N^{nn}\right)\right)^{-1}=\sum_{n=1}^{N}\left(\Lambda\left(\bx_n\right)^{-1}\otimes\mathcal{J}_N^{nn}\right).
\label{equation_F_5}
\end{equation*}

With the above redefinition for the variational approximation, we reduce the time complexity from $\mathcal{O}\left({N}^3{Q}^3\right)$ of the original VEM algorithm to $\mathcal{O}\left({N}{M}^2{Q}^3\right)$, where ${M}$ is the size of the induced covariates ${\underline{X}}$. However, this approximation is still limited: the reduced time complexity is linear to the size of dataset, $N$, but cube to the dimension of responses, $Q$.

\newpage

\section{Auxiliary Figures}\label{secH}

\begin{figure}[!hbt]
\renewcommand{\baselinestretch}{1.0}
\centering
\includegraphics[width=0.24\textwidth] {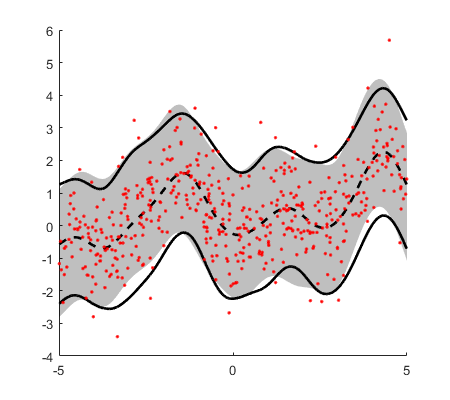}
\includegraphics[width=0.24\textwidth] {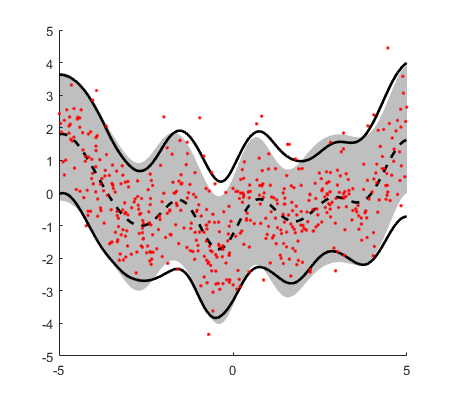}
\includegraphics[width=0.24\textwidth] {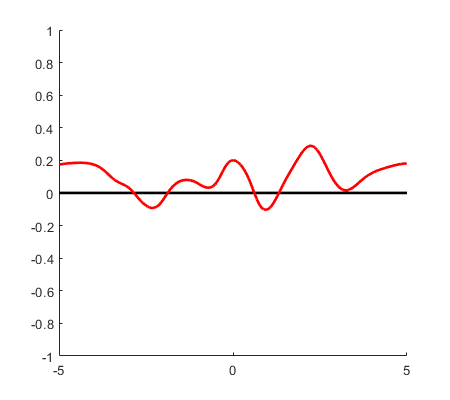}
\includegraphics[width=0.24\textwidth] {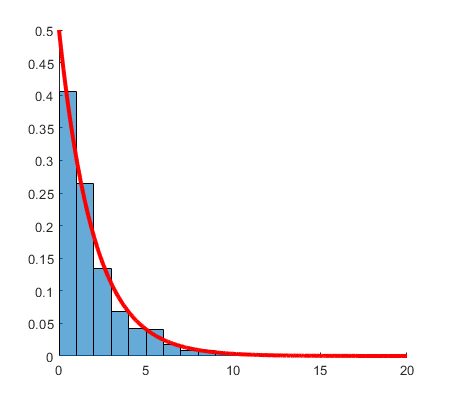}
\caption{(Relevant to Section \ref{sec4_0_1}) The first and second panels are two jointly simulated time series (red dots). Shaded regions are the 95\% confidence bands based on the true generative model, while the black curves are the estimated 95\% confidence bands; the dashed curves are the  medians, respectively. 
The third panel shows the true (black) and estimated (red) residual correlation over covariates. The fourth panel shows the histogram of the square of the standardized residuals from the posterior predictive model overlayed with the $\chi^2{\left(2\right)}$ density (red). 
}
\label{fig_H_4_1_4_1}
\end{figure}

\begin{figure}[!hbt]
\renewcommand{\baselinestretch}{1.0}
\centering
\includegraphics[width=0.9\textwidth] {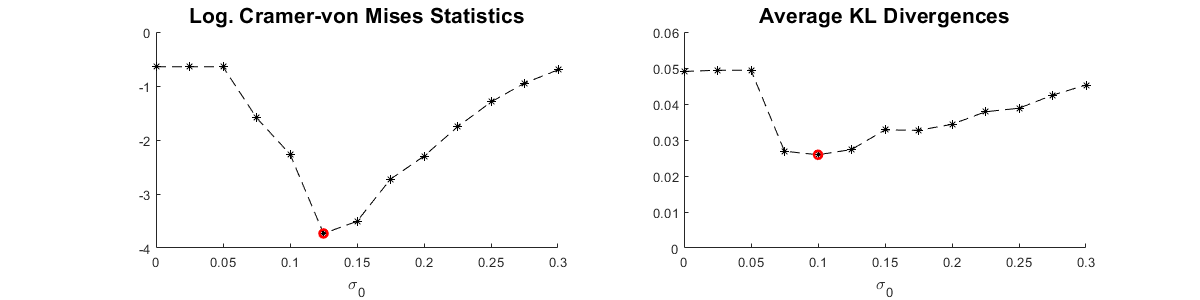}
\caption{(Relevant to Section \ref{sec4_0_1}) Logarithms of the Cramer-von Mises statistics defined in \eqref{equation_4_1_3_17} (left panel) and the average KL divergences from the estimated regression models to the true generative model (right panel) for the heteroscedastic simulated data with outliers, for various $\sigma_0 = 0,0.025,0.05,\cdots,0.275,0.3$. In each panel, the red circle indicates the smallest value.}
\label{fig_4_1_4_2}
\end{figure}

\begin{figure}[!hbt]
\renewcommand{\baselinestretch}{1.0}
\centering
\includegraphics[width=0.9\textwidth] {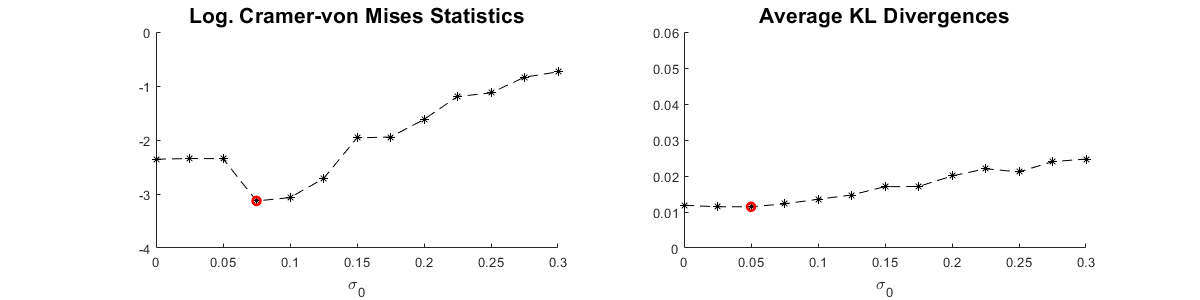}
\caption{(Relevant to Section \ref{sec4_0_1}) Logarithms of the Cramer-von Mises statistics defined in \eqref{equation_4_1_3_17} (left panel) and the average KL divergences from the estimated regression models to the true generative model (right panel) for the heteroscedastic simulated data without outliers, for various $\sigma_0 = 0,0.025,0.05,\cdots,0.275,0.3$. In each panel, the red circle indicates the smallest value.}
\label{fig_H_4_1_4_2}
\end{figure}

\begin{figure}
\renewcommand{\baselinestretch}{1.0}
\centering
\includegraphics[width=0.24\textwidth] {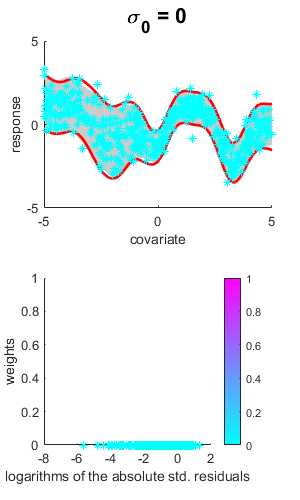}
\includegraphics[width=0.24\textwidth] {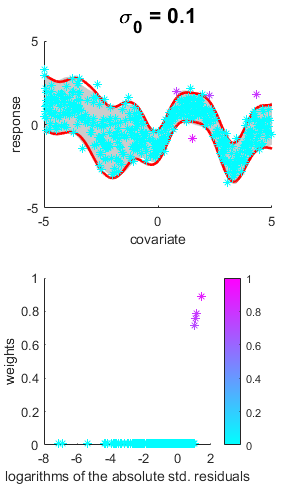}
\includegraphics[width=0.24\textwidth] {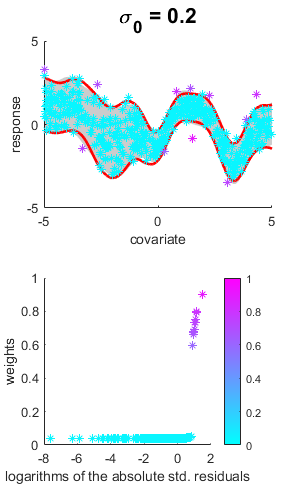}
\includegraphics[width=0.24\textwidth] {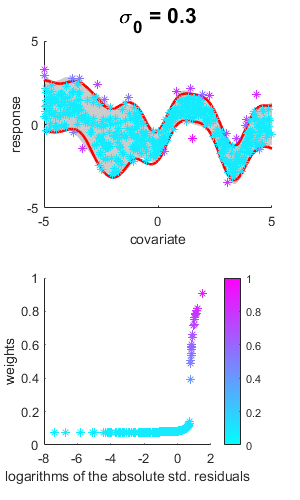}
\caption{(Relevant to Section \ref{sec4_0_1}) A visualization of the estimated weights $\frac{\sigma_0^2}{\hat{\xi}_n^2+\sigma_0^2}$ in the HeGPR-O model for the heteroscedastic simulated data without outliers, for $\sigma_0 = 0,0.1,0.2,0.3$. Colors of the points are corresponding to the estimated weights indicated in the right-side color bar. (Upper) The scatter plot of the simulated responses over the associated covariates. Black regions indicate the 95\% confidence bands of the true generative model and red curves are the estimated 95\% confidence bands by the HeGPR-O model. (Lower) The scatter plot of the estimated weights over the logarithms of the absolute values of the corresponding standardized residuals.}
\label{fig_H_4_1_4_3}
\end{figure}

\begin{figure}
\renewcommand{\baselinestretch}{1}
\centering
\includegraphics[width=0.24\textwidth]{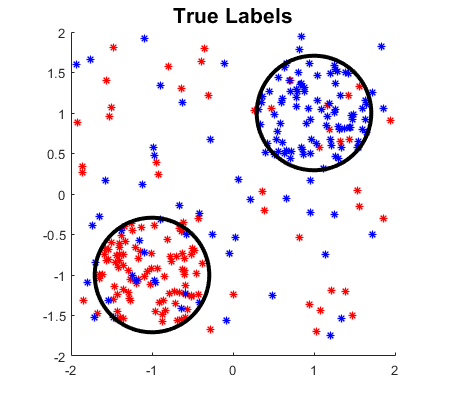}
\includegraphics[width=0.24\textwidth]{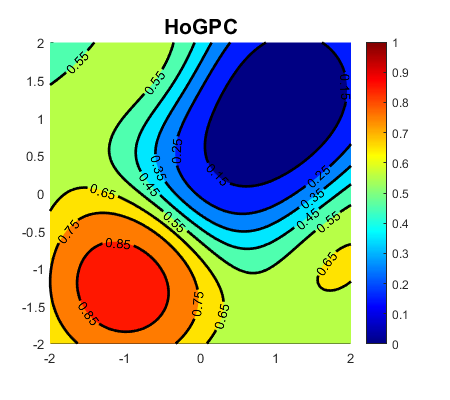}
\includegraphics[width=0.24\textwidth]{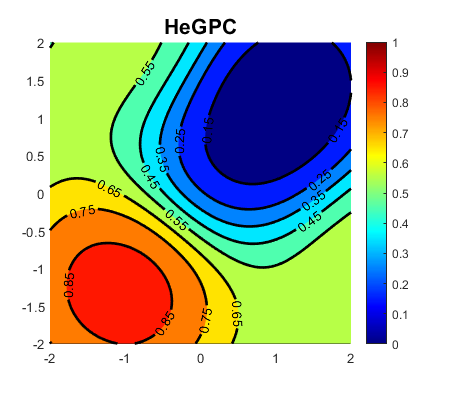}
\includegraphics[width=0.24\textwidth]{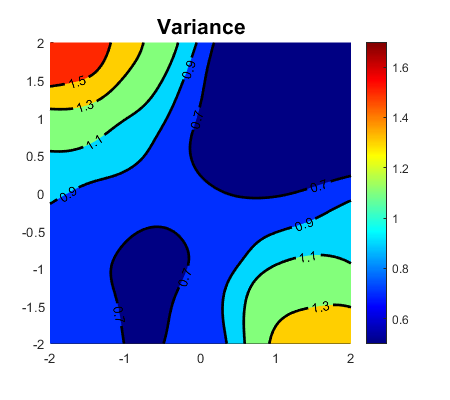}
\caption{(Relevant to Section \ref{sec4_0_2}) Classification results for a simulated dataset on the same simulated dataset used in Figure \ref{fig_4_2_1} but only 10\% of them in 50-50 regions are fed into the algorithm.}
\label{fig_H_4_2_2}
\end{figure}

\begin{figure}
\renewcommand{\baselinestretch}{1}
\includegraphics[width=0.85\textwidth]{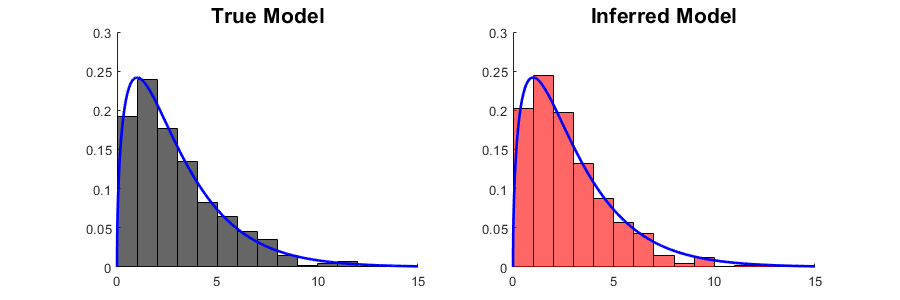}
\caption{(Relevant to Section \ref{sec4_0_3}) Histograms of the chi-squared statistics of the standardized residuals of the true hidden signals. The left panel shows the histogram obtained from the true generative model, and the right one visualizes that from the inferred generative model. Each blue curve is the probability density function of the chi-squared distribution $\chi^2\left(3\right)$.}
\label{fig_4_3_3}
\end{figure}

\begin{figure}
\renewcommand{\baselinestretch}{1}
\includegraphics[width=1.00\textwidth]{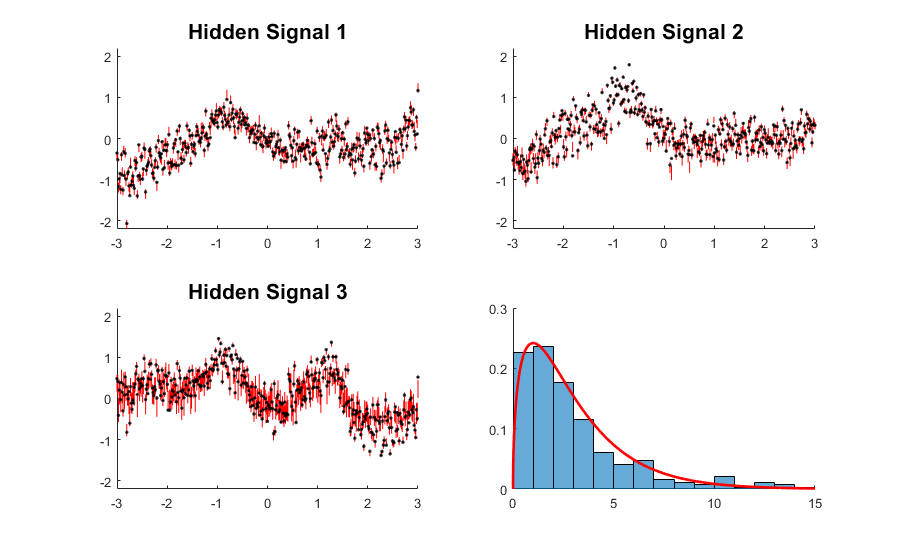}
\caption{(Relevant to Section \ref{sec4_0_3}) The coordinate-wise visualization of the hidden signal estimation and a histogram of the chi-squared statistics of the standardized residuals of the tuples of hidden signals from their posterior distributions compared to the probability density function of the chi-squared distribution $\chi^{2}\left(3\right)$ (red curve). In each of the first three panels, black dots indicate the true hidden signals and red bars are the estimated 95\% confidence intervals from their posterior distributions.}
\label{fig_H_4_3_5}
\end{figure}

\begin{figure}
\renewcommand{\baselinestretch}{1.0}
\centering
\includegraphics[width=0.9\textwidth] {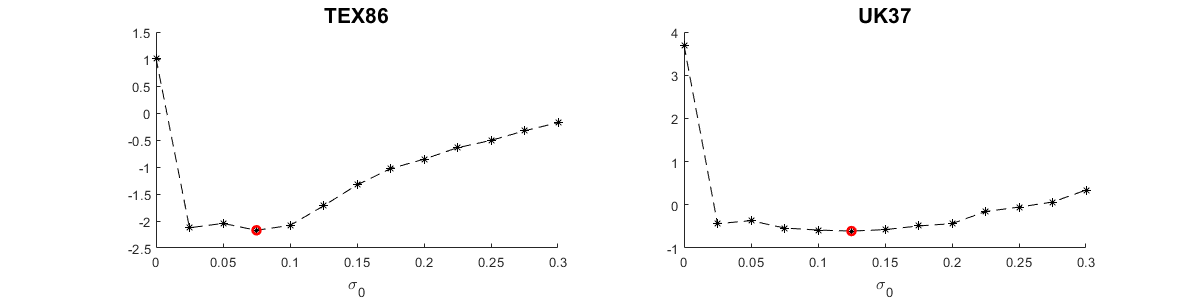}
\caption{(Relevant to Section \ref{sec5_1}) Logarithms of the Cramer-von Mises statistics defined in \eqref{equation_4_1_3_17} for the HeGPR-O modeling on TEX86 (left panel) and UK37 (right panel), for various $\sigma_0 = 0,0.025,0.05,\cdots,0.275,0.3$. In each panel, the red circle indicates the smallest value.}
\label{fig_5_1_1_0}
\end{figure}

\begin{figure}[ht]
\renewcommand{\baselinestretch}{1}
\centering
\begin{tabular}{c|c}
\includegraphics[width=0.5\textwidth]{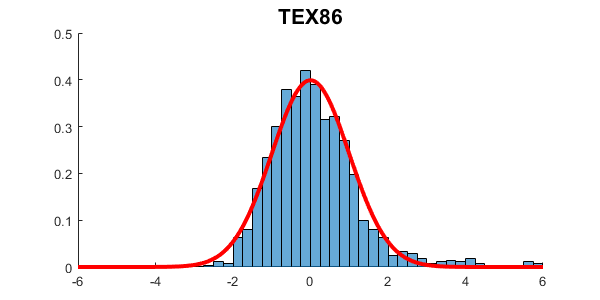} &
\includegraphics[width=0.5\textwidth]{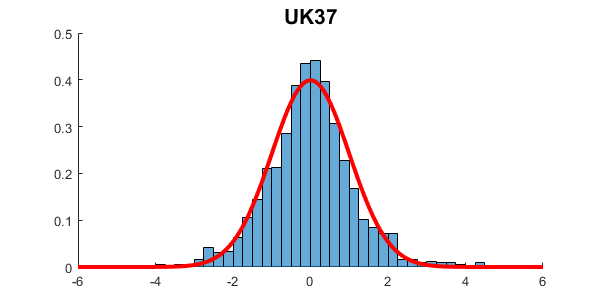} \\
(a) & (b) 
\end{tabular}
\caption{(Relevant to Section \ref{sec5_1}) Histograms of standardized residuals for the logit transformations of TEX\textsubscript{86} and ${\rm{U}}_{37}^{\rm{K}\prime}$. The red curves are pdfs of the standard normal distribution.}
\label{fig_H_5_1_1}
\end{figure}

\begin{figure}
\renewcommand{\baselinestretch}{1}
\centering
\includegraphics[width=0.5\textwidth]{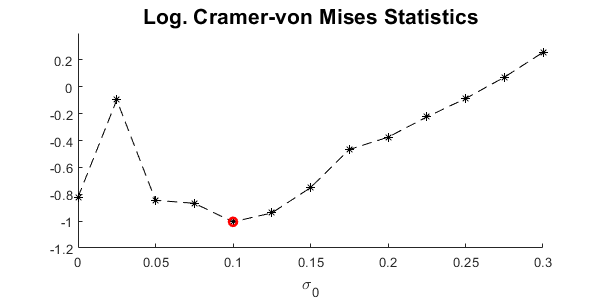}
\caption{(Relevant to Section \ref{sec5_3}) Logarithms of the Cramer-von Mises statistics defined in \eqref{equation_4_1_3_17} for the weekly average temperatures data, for various $\sigma_0 = 0,0.025,0.05,\cdots,0.275,0.3$. The red circle indicates the smallest value.}
\label{fig_5_3_0}
\end{figure}

\begin{figure}
\renewcommand{\baselinestretch}{1}
\centering
\includegraphics[width=1.0\textwidth]{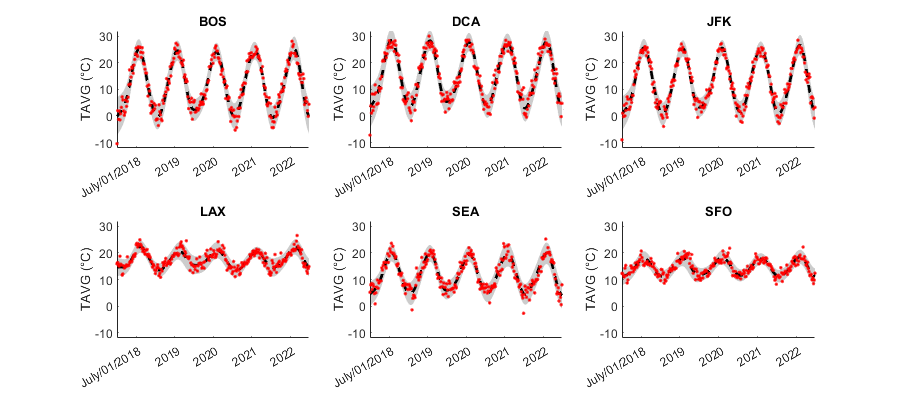}
\caption{(Relevant to Section \ref{sec5_3}) Regression models for the weekly average temperatures of six airports. In each panel, dots indicate the average temperature observations and the shaded region is the inferred 95\% confidence band. Each year in the x-axis is corresponding to the July 1st of that year.}
\label{fig_5_3_1}
\end{figure}

\begin{figure}[ht]
\renewcommand{\baselinestretch}{1}
\centering
\begin{tabular}{c|c}
\includegraphics[width=0.5\textwidth]{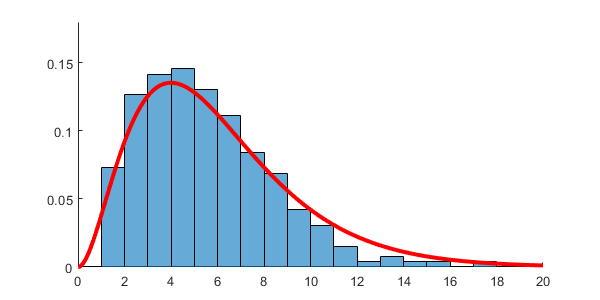} &
\includegraphics[width=0.5\textwidth]{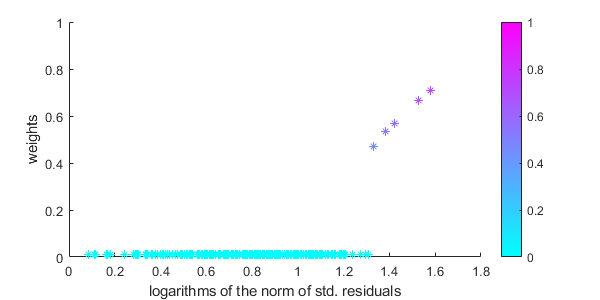} \\
(a) & (b) 
\end{tabular}
\caption{(Relevant to Section \ref{sec5_3}) (a) A histogram of the Chi-squared statistics of the standardized residuals of the tuples of average temperature observations from the regression model. The red curve is the probability density function of the Chi-squared distribution $\chi^2\left(6\right)$. (b) A scatter plot of the pairs of norms of standardized residuals and the corresponding estimated scale parameters.}
\label{fig_H_5_3_3}
\end{figure}

\clearpage
\newpage

\bibliographystyle{unsrtnat}
\bibliography{references}  






\end{document}